\documentclass[jair,twoside,11pt,theapa]{article}
\usepackage{jair, theapa, rawfonts}

\usepackage{amsmath}
\usepackage{graphicx}
\usepackage{caption}
\usepackage{subcaption}
\usepackage{amssymb}

\usepackage{color}
\usepackage[normalem]{ulem}
\newcommand{\sign}{\textrm{sign}}

\jairheading{1}{2022}{1-37}{1/1}{1/1}
\ShortHeadings{Deep Symbol Generation and Rule Learning}{Ahmetoglu, Seker, Piater, Oztop \& Ugur}
\firstpageno{1}

\begin{document}
\title{DeepSym: Deep Symbol Generation and Rule Learning  for Planning from Unsupervised Robot Interaction}

\author{\name Alper Ahmetoglu \email alper.ahmetoglu@boun.edu.tr\\
        \name M. Yunus Seker \email yunus.seker1@boun.edu.tr\\
        \addr Department of Computer Engineering, Bogazici University, Istanbul, Turkey
        \AND
        \name Justus Piater \email justus.piater@uibk.ac.at \\
        \addr Department of Computer Science, Universit\"{a}t Innsbruck, Austria
        \AND
        \name Erhan Oztop \email erhan.oztop@ozyegin.edu.tr \\
        \addr Osaka University, Osaka, Japan / Ozyegin University, Istanbul, Turkey
        \AND
        \name Emre Ugur \email emre.ugur@boun.edu.tr \\
        \addr Department of Computer Engineering, Bogazici University, Istanbul, Turkey}

\maketitle
\begin{abstract}
Symbolic planning and reasoning are powerful tools for robots tackling complex tasks. However, the need to manually design the symbols restrict their applicability, especially for robots that are expected to act in open-ended environments. Therefore symbol formation and rule extraction should be considered part of robot learning, which, when done properly, will offer scalability, flexibility, and robustness. Towards this goal, we propose a novel general method that finds action-grounded, discrete object and effect categories and builds probabilistic rules over them for non-trivial action planning. Our robot interacts with objects using an initial action repertoire that is assumed to be acquired earlier and observes the effects it can create in the environment. To form action-grounded object, effect, and relational categories, we employ a binary bottleneck layer in a predictive, deep encoder-decoder network that takes the image of the scene and the action applied as input, and generates the resulting effects in the scene in pixel coordinates. After learning, the binary latent vector represents action-driven object categories based on the interaction experience of the robot. To distill the knowledge represented by the neural network into rules useful for symbolic reasoning, a decision tree is trained to reproduce its decoder function. Probabilistic rules are extracted from the decision paths of the tree and are represented in the Probabilistic Planning Domain Definition Language (PPDDL), allowing off-the-shelf planners to operate on the knowledge extracted from the sensorimotor experience of the robot. The deployment of the proposed approach for a simulated robotic manipulator enabled the discovery of discrete representations of object properties such as `rollable' and `insertable'. In turn, the use of these representations as symbols allowed the generation of effective plans for achieving goals, such as building towers of the desired height, demonstrating the effectiveness of the approach for multi-step object manipulation. Finally, we demonstrate that the system is not only restricted to the robotics domain by assessing its applicability to the MNIST 8-puzzle domain in which learned symbols allow for the generation of plans that move the empty tile into any given position.
\end{abstract}

\section{Introduction}
\label{sec:intro}
Intelligent robotic systems exploit a diverse range of data representations for control, learning, and reasoning. Interaction with the world requires processing low-level continuous sensorimotor representations whereas abstract reasoning requires the use of high-level symbolic representations. The representational gap between the low-level sensorimotor and high-level symbolic representations has been addressed in AI and robotics, often by using manually designed symbols that are grounded in the low-level sensorimotor experience of the robots interacting with their environment \shortcite{Harnad1990,Taniguchi2019}. However, this approach only works in either controlled environments or limited tasks. Yet, truly intelligent robots are expected to form abstractions \shortcite{Konidaris2019} continually from their interaction with the world and use them on the fly for complex planning and reasoning in novel environments \shortcite{werner1963,callaghan2015}.

In this paper, we address the challenging problem of discovering discrete symbols and unsupervised learning of rules from the low-level interaction experience of a self-exploring robot. For this purpose, we propose a novel deep neural architecture for symbol formation and rule extraction. At the core of our method, the symbols are discovered in the discrete latent space formed by the bottleneck layer of a predictive, deep encoder-decoder network that takes the image of an object and the action applied as the input, and produces the effect generated by the action as the output. Symbols, which are the output of the encoder network, hold information for the effect prediction for a given action. Furthermore, our architecture allows transforming the complete low-level sensorimotor experience into symbolic experience, facilitating direct rule extraction for AI planning. To this end, decision tree models are trained to learn probabilistic rules that are translated to Probabilistic Planning Domain Definition Language (PPDDL; \shortciteR{younes2004ppddl1}) operators that are standard in probabilistic planning. Note that the predicates that appear in the PPDDL operators correspond to the discovered symbols.

In order to realize this framework, we created a setup where a simulated robot manipulator interacts with objects, poking them in different directions and stacking them on top of each other to collect interaction experience for object categorization and rule learning. Our system successfully constructs a latent representation through which object and relational symbols are discovered, which can be interpreted by humans as `rollable', `insertable', `larger-than'. Contrary to symbols generated by systems that disregard actions and effects, our architecture is shown to generate action-effect-regulated symbols that are more effective in abstract reasoning over the actions of the robot and the consequences in the environment. Furthermore, the number of symbols is determined automatically by optimizing the trade-off between prediction capability and bottleneck size. Finally, the system acquired the capability to generate effective plans to achieve goals such as building towers of desired heights from given cubes, balls, and cups using off-the-shelf probabilistic planners. To show the generality of the proposed approach, we also conduct a second set of experiments in a non-robotic domain. To be concrete, we test our approach in the MNIST 8-tile puzzle domain adapted from \shortciteA{asai2017classical}. Our experiments show that the system learns symbols that allow for creating plans to move the empty tile into arbitrary positions.
Our implementation is publicly available\footnote{\texttt{https://github.com/alper111/DeepSym}}.

Our primary contribution is a generic neural solution for mapping raw sensorimotor experience into the symbolic domain. The same architecture can be used to discover object symbols, effect symbols, and object-object relational symbols. The proposed network further allows progressive learning of increasingly complex abstractions, exploiting previously-learned abstractions as inputs. The learned symbols allow abstraction of the interaction of the robot with its environment as a Markov decision process which allows the use of symbolic planning systems for goal satisfaction. In the current study, to show this, we transformed the learned rules into Probabilistic PDDL operators, which allowed probabilistic plan generation and execution achieving goals beyond what was possible with the direct use of the training data.

\section{Related Work}
Bridging the representational gap between the continuous sensorimotor world of a robotic system with the discrete symbols and rules has been a key research goal from the early days of intelligent robotics \shortcite{Kuipers2017,Murphy2000}.
While grounding predefined symbols in the sensorimotor experience of the robot has been widely used for intelligent robot control \shortcite{Klingspor1996,Petrick2008,Mourao2008,Worgotter2009,Kulick2013}, some argue that symbols ``are not formed in isolation'', and ``they are formed in relation to the experience of agents'' \shortcite{Sun2000}. We share this viewpoint that has been investigated in a number of studies. \shortciteA{Pisokas2005} and \shortciteA{Ugur2011RAS} realized systems that clustered low-level sensory experience into categories and performed subsymbolic planning in the continuous perceptual space. While simple planning capability was achieved, the use of continuous prediction and state transition operators limited the use of powerful off-the-shelf symbolic AI planners. In another line of research, \shortciteA{ozturkcu2020high} asked whether there are any symbols formed in a deep RL agent after training the agent for a given task, without imposing any prior on the architecture or the objective.

\shortciteA{mota2019commonsense,riley2019integrating} proposed a hybrid approach to exploit the prior domain knowledge by combining non-monotonic logical reasoning with deep networks. This architecture is a cascade of two models where the first model is the prior domain knowledge encoded as an Answer Set Prolog (ASP) program \shortcite{law2018complexity}, and the second model is a convolutional neural network (CNN). If the ASP program fails to classify an example, it redirects the necessary parts of the input to CNN for further processing. This pipeline results in better accuracy with less computation when compared with CNN classification. Furthermore, given labeled examples about the task, the ASP program can be further extended to include new rules about the environment by using the decision paths of a trained decision tree. These works primarily focus on integrating neural models with common-sense knowledge or domain knowledge to increase performance. Our work is similar to these works in the sense that they also learn previously unknown rules with decision trees from subsymbolic data that would help for the planning. On the other hand, we focus on learning symbols that depend on the action set of the agent.

The bottom-up generation of symbolic structures from the continuous interaction experience of a robot has started to draw attention in robotics \shortcite{Taniguchi2019,Konidaris2019}. \shortciteA{Konidaris2014,Konidaris2015} studied the construction of symbols that are directly used as preconditions and effects of actions for the generation of deterministic and probabilistic plans in 2D agent settings, and \shortciteA{konidaris2018skills} extended the framework into a real-world robot setting. However, these studies use a global state representation, and therefore, symbols learned in an environment cannot be used in a novel environment directly. In follow-up work, \shortciteA{james2019learning} constructs symbols with egocentric representations to allow the transfer of previously learned symbols. These studies train an SVM classifier for each effect cluster to find groundings of precondition symbols. \shortciteA{Ugur-2015-ICRA,Ugur-2015-Humanoids} formed symbols used in plan generation in manipulation using a combination of several ad-hoc machine learning techniques such as clustering with X-means and classification with SVMs. Furthermore, they used hand-crafted features to represent scenes and effects. On the other hand, our proposed architecture simultaneously learns object categories (in the encoder output) and their corresponding effect categories (in the decoder output) without resorting to any clustering techniques on the object or effect space. The object and effect categories automatically emerge as the network with binary bottleneck units minimizes the effect prediction error. Furthermore, deep neural networks allow us to efficiently process high-dimensional image data using convolutional layers. This design approach offers a generic symbol formation engine that runs at the pixel level using deep neural networks. In terms of symbol multiplicity, our approach is more parsimonious, as we do not form symbols for each action as in \shortciteA{Ugur-2015-Humanoids} and \shortciteA{konidaris2018skills}; but instead, use a single decoder network that takes the action as part of the input. To be concrete, for $n$ effect categories and $k$ actions, our system generates $nk$ symbols, whereas the aforementioned approaches generate $n^k$ symbols. Learning a single model for all actions possibly allows internal representations learned for one action to be re-used directly for other actions. Another significant advantage of our model is that it is differentiable and thus can be integrated into gradient-based state-of-the-art machine learning architectures for further tackling more complex problems.

\shortciteA{asai2017classical} realized a similar neural framework where they first train a state autoencoder with discrete latent units, then learn the action precondition-effect mappings. In follow-up work, \shortcite{asai2020learning,asai2022classical} combine these two steps and learn the action mapping together with the state auto-encoder. These works are in the visual domain (for example, 2D puzzles) and achieve visualized plan execution while we focus on robot action planning and execution in the 3D world. Moreover, a critical difference of our method from the aforementioned work is that we learn object symbols by taking into account action and the effects in addition to object features, which facilitates the formation of symbols that are likely to capture object affordances \shortcite{Gibson,Zech17}.

Another line of research focuses on bilevel planning, in which a symbolic plan is complemented by a motion and task planner. \shortciteA{silver2021learning,chitnis2021learning} learn operators for bilevel planning when given parameterized policies for continuous planning. In a follow-up work \shortcite{silver2022learning}, these parameterized policies are learned as well, completing the whole neurosymbolic planning pipeline. While these works fix the state abstractions, \shortciteA{silver2022inventing} also learns new state abstractions that are optimized for planning. In general, these works focus on learning high-level operators for bilevel planning, while we focus on learning symbols from continuous high-dimensional vectors. Another similar work \shortcite{yuan2022sornet} trains a network that outputs relations between objects from RGB images given objects' canonical images.

\section{Problem Formulation}
\label{sec:problem}
In this work, we refer to symbols as discrete low-dimensional vectors extracted from deep neural networks for the current state and used to predict the observed effect of specific actions. More formally, a symbol $\mathbf{z} \in \mathcal{Z}$ is a discrete representation that represents a subset $\mathcal{P}$ of a continuous high-dimensional space $\mathbb{R}^n$ (e.g., the state-space, or the effect-space). The symbol-space $\mathcal{Z}$ can be defined as a set of $k$-dimensional boolean vectors $\mathcal{Z}=\mathbb{B}^m=\{0, 1\}^m$, or as a set of atoms $\mathcal{Z}=\{z_1, z_2, \dots, z_m\}$. The important condition here is that the symbol-space should be finite, and its cardinality $|\mathcal{Z}|$ should preferably be small. In general, the symbol learning problem refers to finding the mapping $f: \mathbb{R}^n \rightarrow \mathcal{Z}$, which would allow us to do logical reasoning in the symbolic domain.

Given a set of discrete actions $\mathcal{A}=\{a_1, a_2, ..., a_k\}$, continuous object (or state) space $\mathbb{R}^n$, and continuous effect space $\mathbb{R}^m$, we are interested in learning an encoder function $f: \mathbb{R}^n \rightarrow \mathcal{Z}$ and a decoder function $g: \mathcal{Z} \times \mathcal{A} \rightarrow \mathbb{R}^m$ from samples $\{\mathbf{o}^{(i)}, a^{(i)}, \mathbf{e}^{(i)}\}_{i=1}^N$ collected by interacting with the environment. Essentially, the encoder outputs symbol $\mathbf{z}$ given the object state $\mathbf{o} \in \mathbb{R}^n$, and the decoder outputs effect $\mathbf{e} \in \mathbb{R}^m$ for symbol $\mathbf{z}$ and action $a$. After learning the encoder and the decoder function by iteratively optimizing an objective (which will be discussed in Section \ref{sec:methods}), $\mathbf{z}$ corresponds to an object symbol, and $c$ corresponds to an effect symbol that has the grounding $\mathbf{e}=g(\mathbf{z}, a)$ (note that $c$ is an atom while $\mathbf{e}$ is a continuous vector). Once we have such symbols, we can construct a high-level plan in the symbolic space by transforming the environment to a probabilistic PDDL domain defined over the symbols, and then use state-of-the-art off-the-shelf planners to find an action sequence that arrives at the desired goal state.

The experiments reported in this paper involve two environments from different domains, namely, a tabletop robotic manipulation environment and MNIST 8-puzzle environment adapted from \shortcite{asai2017classical}. The former is an embodied robotic environment in which symbols that emerge depend on the actions executed by a robotic arm and their corresponding effects. In the MNIST 8-puzzle environment, an agent without an embodiment executes actions and observes the corresponding effects as the visual change in the environment. Symbols are learned with respect to these actions and visual effects.

For simplification, we make the following assumptions in the tabletop manipulation environment:
\begin{itemize}
    \item The agent is assumed to have a small number of actions, such as poking and stacking an object. Such an action repertoire can be autonomously acquired through a developmental progression as in \shortcite{Ugur2012} or obtained through learning from demonstration and reinforcement learning \shortcite<e.g.,>{Seker2019,Akbulut2020}.

    \item The agent is equipped with image processing capability to detect the objects in the camera image and also calculate their pixel coordinates. Furthermore, using the same object tracking method, the agent can take cropped images as input. In the tabletop setup, we realized this with a simple algorithm as the background is uncluttered. In a real-world scenario, state-of-the-art computer vision techniques can be used to detect and track objects in the 3D world.
\end{itemize}
In the MNIST 8-puzzle environment, the only assumption is that the agent has access to the action repertoire (e.g., `slide-left', `slide-down'), which it can execute to see the effects of its actions.

\begin{figure*}[t]
\centering
\includegraphics[width=\textwidth]{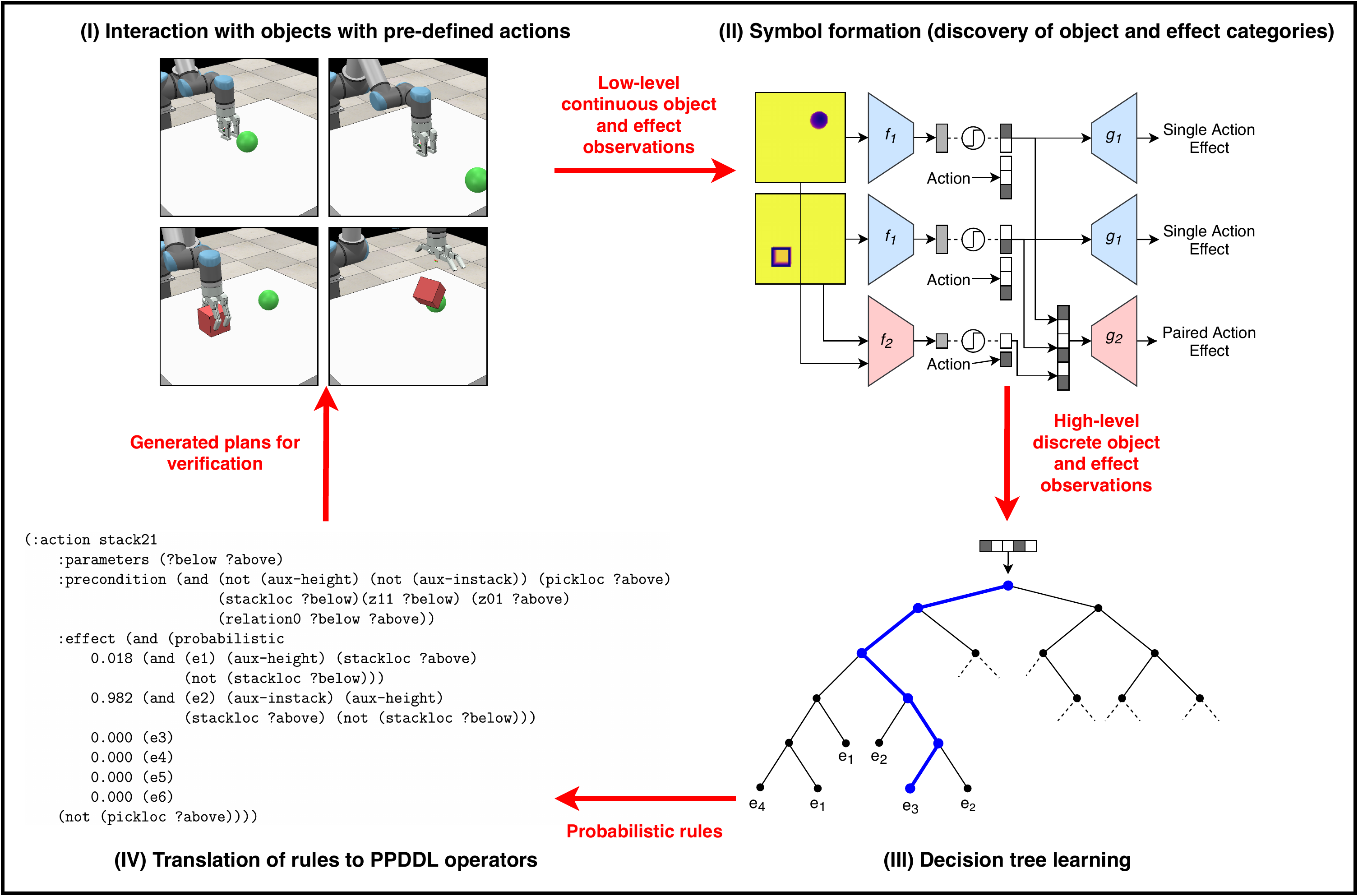}
\caption{General system overview of rule generation and refinement.}
\label{fig:pipeline}
\end{figure*}

\section{Methods}
\label{sec:methods}

Figure~\ref{fig:pipeline} provides the overall learning architecture of our proposed system in the robotic manipulation environment; the application of the architecture to the MNIST 8-puzzle domain is given in Section~\ref{sec:8puzzle}. In the environment interaction phase (I), the robot chooses an action from its action repertoire $a \in \mathcal{A}=\{a_1, a_2, \dots, a_k\}$, observes the object state $\mathbf{o}$, executes the action, and records the resulting effect $\mathbf{e}$.

Using the interaction experience $\{\mathbf{o}^{(i)}, a^{(i)}, \mathbf{e}^{(i)}\}_{i=1}^{N}$, the symbol formation is achieved in (II). To this end, a deep neural network model with two parts is trained to predict $\mathbf{e}$ given $\mathbf{o}$ and $a$. The first part is the encoder network, $f(\mathbf{o})$, which creates a \emph{binary} latent vector $\mathbf{z}$ given the depth image of the object, $\mathbf{o}$. The second part, the decoder network $g(\mathbf{z}, a)$, predicts the effect $\mathbf{e}$ when action $a$ is executed on state $\mathbf{o}$ that has the latent representation $\mathbf{z}$. As the network tries to predict effects, symbolic representations are created by the encoder network that can be treated as object categories regulated by the corresponding action-effect experience.

The continuous interaction experience $\{\mathbf{o}^{(i)}, a^{(i)}, \mathbf{e}^{(i)}\}_{i=1}^{N}$ is transformed into the symbolic experience $\{\mathbf{z}^{(i)}, a^{(i)}, c^{(i)}\}_{i=1}^{N}$ using the discovered categories, and then the symbolic experience is used to distill a decision tree to predict effects given object categories and actions in (III). The reason to use a decision tree is that we can represent any statement in propositional logic with decision trees \shortcite[Ch. 19.3]{aima} and we can convert rules of the environment into logical statements that encode pre- and post-conditions of actions on the objects.

Finally, these statements are represented in PPDDL, which allows one to make plans in a probabilistic environment in (IV). Lastly, plans are executed to validate the learned symbols and rules. In the following sections, we describe these parts in detail.

\subsection{Exploration with the Environment}
\label{subsec:exploration}

A manipulator robot with a gripper and a depth camera is used to explore the environment and monitor the changes (Figure~\ref{fig:exp_setup}). The robot is initialized with a fixed set of actions $\mathcal{A}=\{a_1, a_2, \dots, a_k\}$ through which it interacts with the objects in its workspace. Forward, side, and top poking actions are used to poke objects from different sides (Figure~\ref{fig:actions}, top). The stacking action is used to release one object on top of another object (Figure~\ref{fig:actions}, bottom). These actions are encoded with one-hot encoding. On the perception side, each detected object is represented with its top-down depth image. The generated change, on the other hand, is represented by the positional offset of the acted object in pixel coordinates together with the force change sensed at the wrist joint of the robot. In single-object interactions, the robot observes and stores the initial state as the object-centered, top-down depth image of the object, and the effect as the change in object position and force sensor readings:
\begin{equation}
    \mathbf{e}_{\textnormal{single}} = (\Delta x, \Delta y, \Delta d, \Delta F)
\end{equation}
where $\Delta x$ and $\Delta y$ are the changes in $x$-axis and $y$-axis in pixel coordinates, respectively, $\Delta d$ is the change in depth, and $\Delta F$ is the change in force. In paired-object interactions, the robot observes and stores the initial state as the combination of two object-centered depth images $(o_1, o_2)$, and the effect as the change in position of both objects:
\begin{equation}
    \mathbf{e}_{\textnormal{paired}} = (\Delta x_1, \Delta y_1, \Delta d_1, \Delta x_2, \Delta y_2, \Delta d_2)
\end{equation}
where $\Delta x_1$, $\Delta y_1$, $\Delta d_1$ refer to the displacement of the first object, and $\Delta x_2$, $\Delta y_2$, $\Delta d_2$ refer to the displacement of the second object.

\subsection{Symbol Discovery with Deep Networks}
\label{subsec:deepnets}

\begin{figure}
\centering
\includegraphics[width=\linewidth]{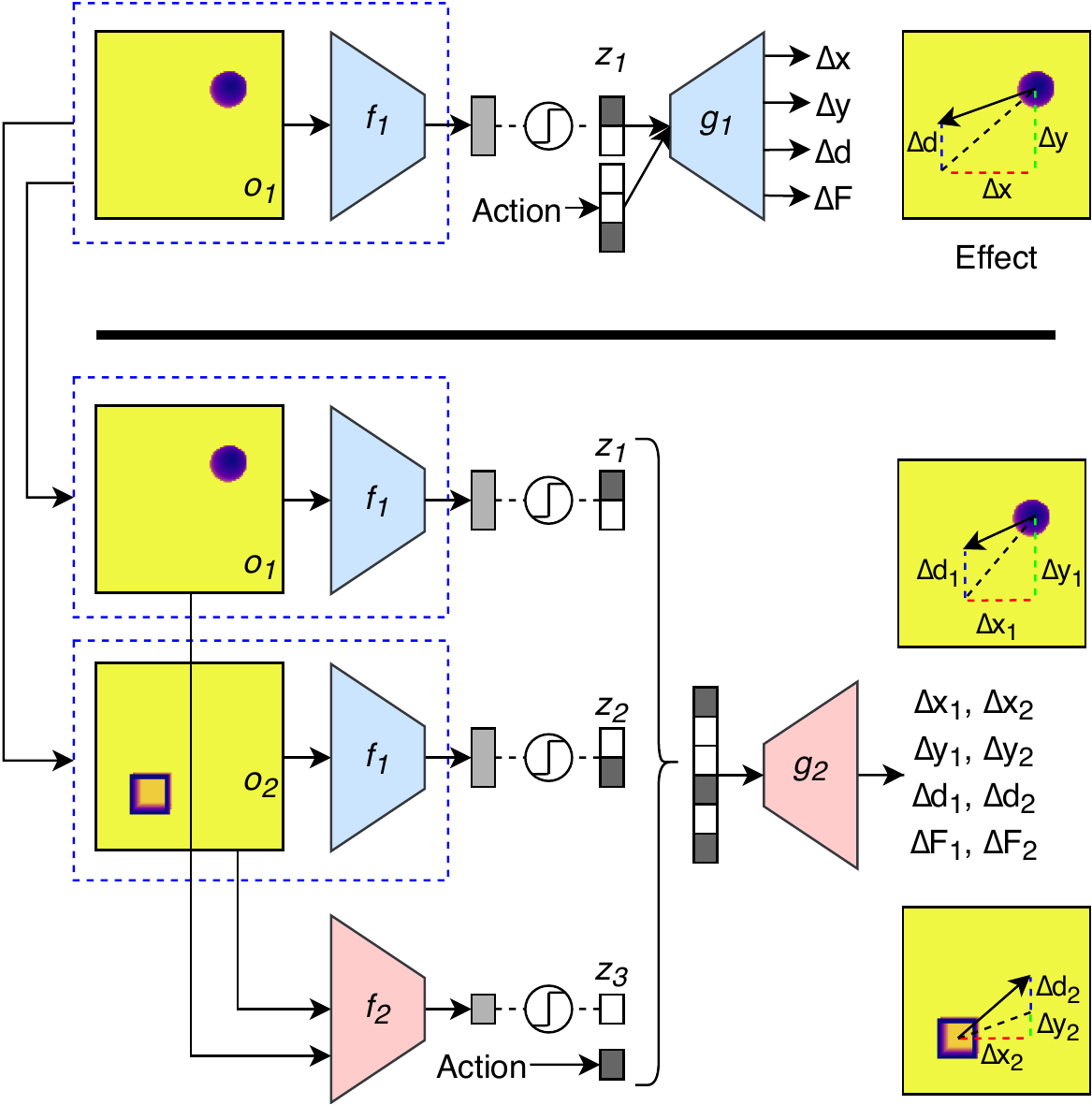}
\caption{Network architectures for single-object interactions (top) and for paired-object interactions (bottom).}
\label{fig:network}
\end{figure}

The main objective of the network is to discover symbols, i.e., object and effect categories, that are effective in abstract reasoning about the consequences of robot actions. In other words, the object categories, together with robot actions, should give the ability to predict the effect categories. To achieve this, we propose a special neural network structure which is composed of two parts: an encoder $f(\mathbf{o})$ to predict $\mathbf{z}$ which is the object category, and a decoder $g(\mathbf{z}, a)$ to predict $\mathbf{e}$ (Figure~\ref{fig:network}, top). This is an encoder-decoder design that has been shown to be quite successful in many different applications \shortcite{hinton2006reducing,kingma2013auto,sutskever2014sequence,devlin2018bert}. The binary bottleneck layer forces the network to learn low-dimensional symbolic representations that are useful for predicting the generated effect of actions. As the input is a top-down depth image, the encoder is a convolutional neural network with the Gumbel-Sigmoid (GS) function \shortcite{maddison2016concrete,jang2016categorical} as the last-layer activation function (where the error back-propagation is handled with the reparameterization trick; \shortciteR{kingma2013auto}). We also experimented with the $\sign(x)$ function using straight-through estimators (STE; \shortciteR{bengio2013estimating}) and found that GS has a lower variance. Results with STE are given in Appendix \ref{app:ste}. Using GS activation of the bottleneck neurons, the continuous representation is directly transformed into a discrete category.
The decoder part is realized as a multi-layer perceptron (MLP). The category $\mathbf{z}$ of the object $\mathbf{o}$ concatenated with the one-hot vector of action $a$ is given to the decoder as input. The decoder predicts the effect $\mathbf{e}$ expected to be observed on object state $\mathbf{o}$ via action $a$. The network minimizes the following objective:
\begin{gather}
\label{eq:mse}
\mathcal{L} = \sum_{i=1}^N \frac{1}{2}\left(g(f(\mathbf{o}^{(i)}), a^{(i)})  - \mathbf{e}^{(i)}\right)^2
\end{gather}

\begin{figure}
    \centering
    \begin{subfigure}[b]{0.49\textwidth}
        \centering
        \includegraphics[width=\textwidth]{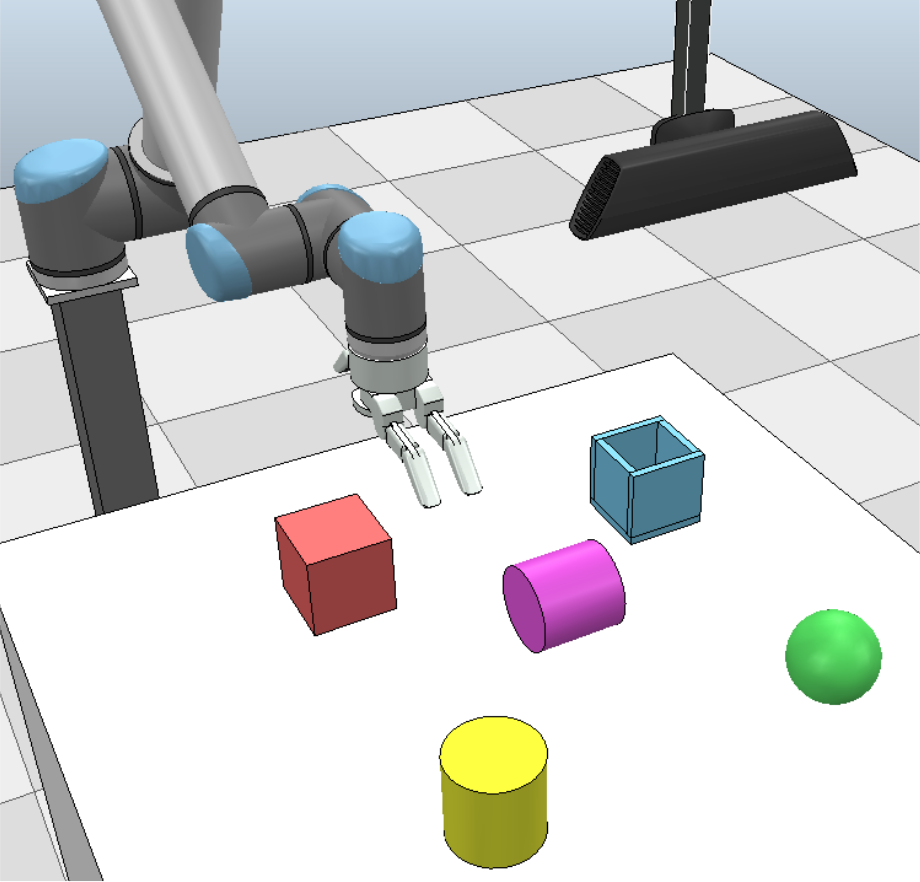}
        \caption{Experiment setup.}
        \label{fig:tabletop}
    \end{subfigure}
    \hfill
    \begin{subfigure}[b]{0.47\textwidth}
        \centering
        \includegraphics[width=\textwidth]{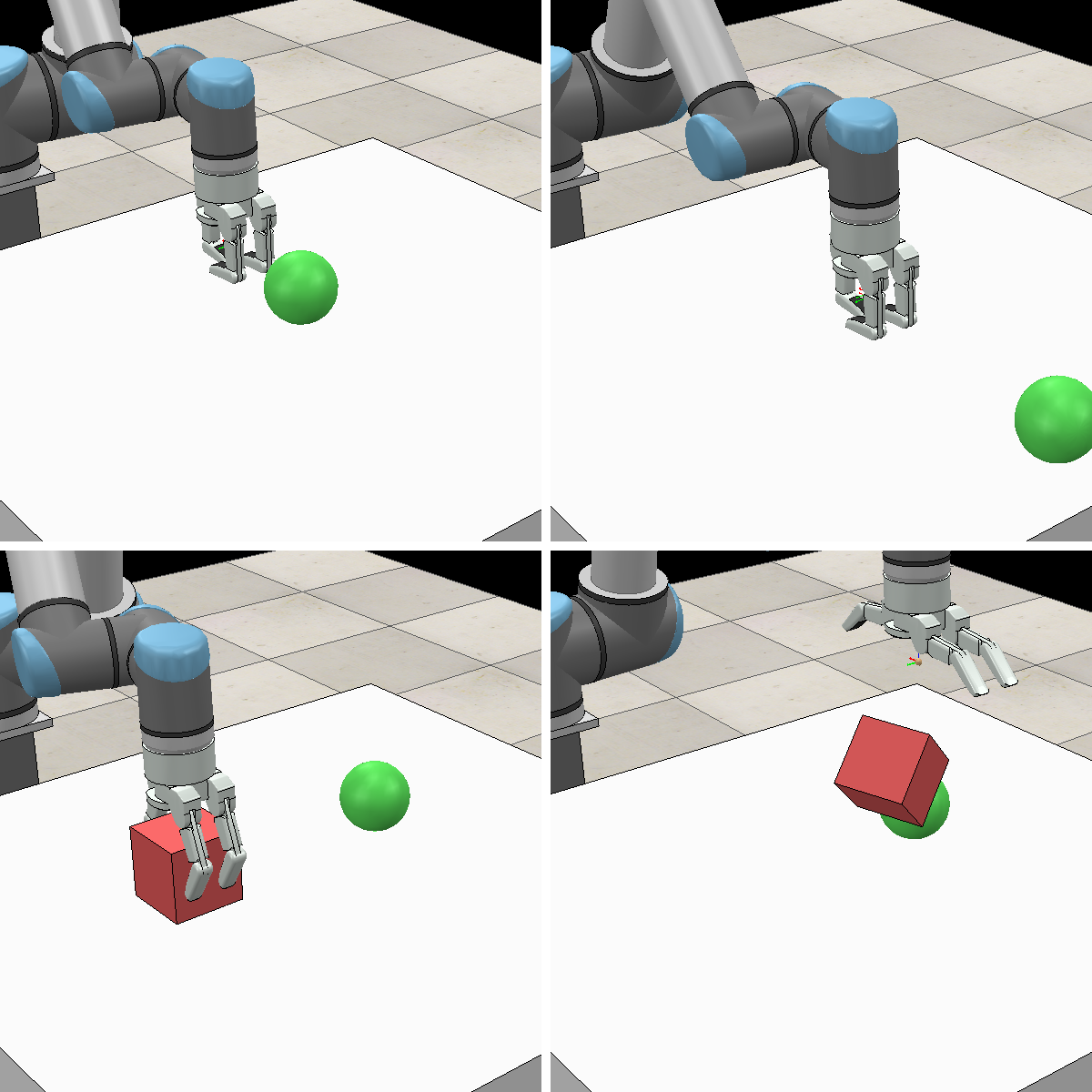}
        \caption{Available actions.}
        \label{fig:actions}
    \end{subfigure}
    \caption{A simulated UR10 robot arm and BarrettHand grasper are used for manipulation; a Kinect sensor is used for perception. Five types of objects are shown on the table.}
    \label{fig:exp_setup}
\end{figure}

This architecture effectively creates high-level symbolic categories of objects that encapsulate the effects of executed actions. One important advantage is that the model does not need hand-engineered object features and object clusters for finding object symbols, contrary to previous studies, since the system learns \emph{discrete} categories directly to optimize the effect prediction performance. Moreover, as the bottleneck layer is discrete, the possible decoder outputs 
$\mathbf{e}=g(\mathbf{z}, a)$ form a finite set 
$\mathcal{E}=\{\mathbf{e}_1, \mathbf{e}_2 \dots\}$ which can be denoted by atoms $\mathcal{C}=\{c_1, c_2, \dots\}$. The learned object categories also serve as input for the discovery of new categories with new interaction experience such as actions applied on pairs of objects such as stacking an object $\mathbf{o}_1$ on top of another object $\mathbf{o}_2$ (Figure~\ref{fig:actions}, bottom). The same deep network structure is used to extract the corresponding symbols with a slight modification to incorporate previously learned knowledge (Figure~\ref{fig:network}, bottom). Here, an encoder $f_2$ takes the depth images of the objects and produces a binary latent vector $\mathbf{z}_3$. As the important point here, the single object symbols ($\mathbf{z}_1$ and $\mathbf{z}_2$) computed by the $f_1$ encoder are also added to the network as input together with the action information. The idea is that we can use previously-acquired symbols to encode new information more compactly, thus allowing a progressive increment of symbols. Note that the encoder $f_1$ is frozen at this second stage of the training. The encoder $f_1$ provides some interaction related information about objects and let the encoder $f_2$ focus and learn properties and relations \emph{between} the objects.

\paragraph{Number of symbols} is automatically set by selecting the number of bottleneck neurons using a hyperparameter search procedure. To limit the number of rules and predicates, this procedure aims to find the minimum number of symbols that provide competitive performance in prediction. Starting from one unit, we record the mean and the standard deviation of mean square errors (MSE) of multiple runs. We increase the number of units until there is no significant drop in the prediction error. MSE curves are reported in Appendix \ref{app:network}.

\subsection{Extracting Symbolic Rules}
\label{subsec:rules}
In the third part of the pipeline, a decision tree is trained to predict the effect $c$ of the stack action $a$ given high-level single ($\mathbf{z}_1$ and $\mathbf{z}_2$) and paired ($\mathbf{z}_3$) object categories (i.e., the dataset is \{$[\mathbf{z}_1^{(i)}; \mathbf{z}_2^{(i)}; \mathbf{z}_3^{(i)}; a^{(i)}], c^{(i)}\}_{i=1}^{N}$). Here, the aim is to extract the probabilistic rules of the environment by converting the decision rules on the paths of the tree into logical statements, which ultimately enables probabilistic planning.
Each path from the root node to a leaf node in the decision tree stores the required set of predicates $\{p_1=(z_3 < 0.5), p_2=(z_2 > 0.5), \dots\}$ represented by discovered single and paired-object categories (in the internal nodes) in order to achieve the effect category $c$ (in the leaves). In other words, each path corresponds to a set of preconditions in order to reach a different effect. As the decision rules at each node in a path $\mathcal{P}$ are in conjunction ($p_1 \land p_2 \land \dots \land p_k$), and these paths are in disjunction ($\mathcal{P}_1 \lor \mathcal{P}_2 \lor \dots \lor \mathcal{P}_m$), the tree represents a statement in disjunctive normal form. Thus, any statement in propositional logic can be represented as a decision tree \shortcite[Ch. 19.3]{aima}. The class probabilities at a leaf (the fraction of samples) correspond to probabilities of observing different effects for the same set of preconditions. Therefore, each path is directly converted to a different rule in probabilistic PDDL. While training the decision tree, the minimum number of samples required for a node to be a leaf node is empirically set to 100 samples. The extracted rules are only limited to predicting the effects of an action. In this way, the agent is not expected to learn representations (and consequently rules) unrelated to its embodiment and actions. For example, in our tabletop environment, the robot cannot differentiate cubes from vertical cylinders as different categories since they respond similarly to similar actions even though their visual appearances differ.

Our motivation to construct PPDDL descriptions is to use probabilistic AI planners to efficiently make plans and execute them. PPDDL is composed of a domain description and a problem definition. In the domain description, there are predicates and actions. Predicates represent boolean values that can be activated or deactivated. Each action has a precondition, which is a set of predicates that needs to be satisfied, and an effect, which activates/deactivates other predicates. The domain description is generated from the list of rules. In the problem definition, the initial state of the world is encoded along with the goal to be satisfied. To encode the initial state, the robot perceives the current environment and sets the truth values of the predicates for the existing categories. The planner finds the sequence of actions to satisfy the predicates given in the goal description starting from the initial state using the actions defined in the domain description.

\section{Robot Experiments}
\label{sec:tabletop_exp}
In the following experiments, we aim to answer the following questions to evaluate the proposed method:
\begin{enumerate}
    \item Do the learned symbols hold any high-level meaning?
    \item Are the learned symbols effective for symbolic planning?
\end{enumerate}
\noindent We compare our method with two alternative baselines:
\begin{enumerate}
    \item An autoencoder with discrete activations where symbols are learned directly from passively observed states, independent from actions and effects.
    \item An encoder-decoder network with continuous activations, followed by clustering in the latent space.
\end{enumerate}
\noindent Regarding the first question, we evaluate the methods based on their performance in differentiating object categories. For the second question, we evaluate the planning performance of different methods.

\subsection{Experiment Setup}
\label{subsec:exp_setup}
\paragraph{Interactions:} We adopted the robotic setup, including the action and object sets used, from \shortciteA{Ugur-2015-ICRA} who showed effective skill transfer from the simulator to real-world, involving actions with 3-fingered prehension. The experiments are performed in CoppeliaSim VREP simulator \shortcite{coppeliaSim} where a six-degrees-of-freedom UR10 \shortcite{ur10} robot arm and a Barrett Hand system \shortcite{townsend2000barretthand} interacts with the objects on the table, and a top-down facing Kinect sensor is used for environment perception (Figure~\ref{fig:exp_setup}). The objects used in the experiments include rectangular cups, horizontally and vertically placed cylinders, spheres, and cubes. For each object type, ten different objects with varying diameters/edge lengths in the range of 10 to 20 cm are included in the object dataset for interaction.

\paragraph{Perception:} Before each action execution a top-down depth image ($128 \times 128$ pixels) of the scene is captured. Objects are placed at different reachable locations on the table during the interactions to ensure the network is invariant with the perspective. Pixels of the images are normalized globally to increase the convergence speed of stochastic gradient descent \shortcite{lecun2012efficient}. Objects in the image are detected with a simple procedure by finding the point with minimum depth and cropping the area of $42 \times 42$ pixels centered around it. This procedure yields object-centered representations for the objects used in the current study but preserves the perspective distortion due to varying locations of the objects and fixed sensor position.

\paragraph{Encoder-decoder network:} The encoder network (Figure~\ref{fig:network}) consists of four blocks each containing two convolutional layers that are followed by batch normalization \shortcite{ioffe2015batch} and ReLU activation. The number of filters in these blocks are 32, 64, 128, and 256. The last layer consists of two hidden units with a GS activation.
The decoder network is a two-layer MLP with 32 hidden units. Further details of these networks can be found in Appendix \ref{app:network}.
 
\subsection{Discovered Object Categories}
\label{subsec:cat}

\begin{table}
 \centering
 \begin{tabular}{l|cccc}
 \hline
 % this is deepsym + gs
 \multicolumn{5}{c}{DeepSym} \\
 \hline
 Category & (0, 0) & (0, 1) & (1, 0) & (1, 1) \\
 \hline
 Sphere & \textbf{99.9 $\pm$ 0.2} & 0.1 $\pm$ 0.2 & 0.0 $\pm$ 0.0 & 0.0 $\pm$ 0.0 \\
 Cube & 0.0 $\pm$ 0.0 & \textbf{99.9 $\pm$ 0.2} & 0.0 $\pm$ 0.0 & 0.1 $\pm$ 0.2 \\
 Vertical Cylinder & 0.0 $\pm$ 0.0 & \textbf{99.9 $\pm$ 0.2} & 0.1 $\pm$ 0.2 & 0.0 $\pm$ 0.0 \\
 Horizontal Cylinder & 0.4 $\pm$ 0.9 & 3.1 $\pm$ 5.5 & \textbf{93.0 $\pm$ 4.7} & 3.4 $\pm$ 3.8 \\
 Cup & 0.0 $\pm$ 0.0 & 0.0 $\pm$ 0.0 & 0.0 $\pm$ 0.0 & \textbf{100.0 $\pm$ 0.0} \\
 \hline
 % this is autoencoder + gs
\multicolumn{5}{c}{Autoencoder (OBO)} \\
 \hline
 Category & (0, 0) & (0, 1) & (1, 0) & (1, 1) \\
 \hline
 Sphere & \textbf{60.6 $\pm$ 1.7} & 15.7 $\pm$ 5.0 & 15.0 $\pm$ 4.5 & 8.8 $\pm$ 3.2 \\
 Cube & \textbf{37.4 $\pm$ 2.3} & 22.9 $\pm$ 3.1 & 19.3 $\pm$ 2.4 & 20.4 $\pm$ 4.3 \\
 Vertical Cylinder & \textbf{44.0 $\pm$ 2.9} & 23.2 $\pm$ 5.7 & 19.4 $\pm$ 7.2 & 13.4 $\pm$ 3.1 \\
 Horizontal Cylinder & \textbf{44.7 $\pm$ 2.9} & 20.0 $\pm$ 4.9 & 18.6 $\pm$ 3.7 & 16.7 $\pm$ 4.5 \\
 Cup & \textbf{86.5 $\pm$ 2.0} & 4.3 $\pm$ 2.1 & 6.4 $\pm$ 2.3 & 2.8 $\pm$ 3.7 \\
 \hline
 \multicolumn{5}{c}{Continuous bottleneck + clustering (OCEC)} \\
 \hline
 Category & (0, 0) & (0, 1) & (1, 0) & (1, 1) \\
 \hline
 Sphere & \textbf{94.2 $\pm$ 9.4} & 4.0 $\pm$ 8.4 & 1.8 $\pm$ 5.5 & 0.0 $\pm$ 0.0\\
 Cube & 0.0 $\pm$ 0.0 & \textbf{99.0 $\pm$ 3.2} & 0.0 $\pm$ 0.0 & 1.0 $\pm$ 3.2\\
 Vertical Cylinder & 0.0 $\pm$ 0.0 & \textbf{98.5 $\pm$ 4.7} & 0.0 $\pm$ 0.0 & 1.5 $\pm$ 4.7\\
 Horizontal Cylinder & 28.6 $\pm$ 29.1 & 0.0 $\pm$ 0.0 & \textbf{63.4 $\pm$ 26.1} & 8.0 $\pm$ 16.9\\
 Cup & 0.0 $\pm$ 0.0 & 13.8 $\pm$ 32.5 & 0.0 $\pm$ 0.0 & \textbf{86.2 $\pm$ 32.5}\\
 \hline
 \end{tabular}
 \caption{The relative assignment frequencies of objects to different symbolic categories. Here, objects vary in their sizes and initial positions. The mean and the standard deviation of 10 runs are reported. For ease of understanding, we name columns so that the category where spheres are mostly placed is renamed to (0, 0), the category where cubes are mostly placed is renamed to (0, 1), and so on. The naming convention also allows us to take an average across different runs.}
 \label{tab:singlecats}
\end{table}

\begin{figure}
 \centering
 \includegraphics[width=\linewidth]{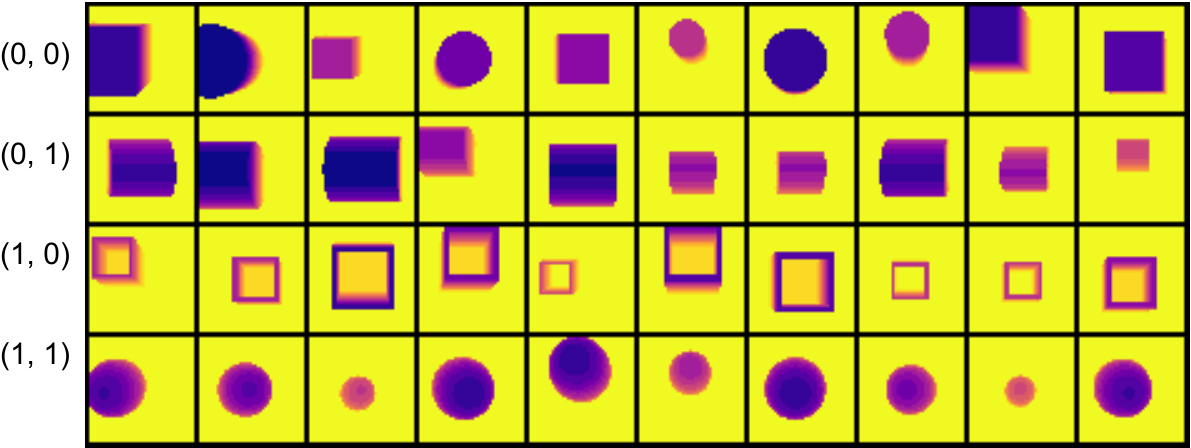}
 \caption{Example depth images as inputs to the encoder network $f_1$.}
 \label{fig:missed}
\end{figure}

Based on the hyperparameter optimization procedure, the number of binary activation neurons in the bottleneck layer is automatically set to 2; therefore, the system found $2^2=4$ object categories. How different object types (unknown to the robot) are represented by the discovered object categories is analyzed and provided in Table~\ref{tab:singlecats}.
In general, different types of objects were coded into different categories except that cube and vertical cylinder share the same category even though their depth images differ. This is due to our action and effect regulated categorization: cubes and vertical cylinders behave the same under all available single-object actions of the robot. Although the depth images of the same type of objects with different sizes differ significantly, this information is not reflected in the categories because the size of the objects does not have a significant influence on the consequences of the current actions. The categories can be interpreted as `pushable'; `rollable in single direction'; `pushable and insertable'; and `rollable in all directions', respectively. Examples from each category are shown in Figure~\ref{fig:missed}.

As a baseline for comparison, we trained
\begin{enumerate}
    \item An autoencoder with a binary hidden layer using Gumbel-Sigmoid to reconstruct the depth images of objects (inputs to $f_1$) instead of effects. This approach is similar to \shortciteA{asai2017classical}. Let us refer to this approach as Object-Binary-Object (OBO).
    \item Our proposed encoder-decoder architecture with the binary bottleneck layer replaced with a usual continuous layer that is applied $k$-means clustering ($k=4$) after learning. Let us call this approach Object-Continuous-Effect followed by Clustering (OCEC).
\end{enumerate}
The results are shown in Table~\ref{tab:singlecats}. For the autoencoder network (i.e., OBO), we see that objects are collapsed primarily into one category. The robot is expected to predict the consequences of its actions using these categories, and as shown, these categories are not distinctive to help such prediction. With this, we verified the advantage of extracting the symbols from the interaction experience of the robot that includes object-action-effect information, i.e., from an object encoder - effect decoder network, rather than searching the symbols in \emph{passively-observed} static features.

OCEC gave better results compared to OBO since the bottleneck layer in OCEC \emph{does} include information from the effect space because of the predictive training similar to our proposed model. However, the latent codes in the bottleneck layer of OCEC might not be distributed locally, making clustering harder. When this is the case, we need more complicated clustering algorithms such as spectral clustering to cluster the latent space accurately. For example, in Table \ref{tab:singlecats}, we see that OCEC is more biased toward misclassifying cups as the stable category and the horizontal cylinders as spheres. When we take an average over all objects, our method predicts objects in the correct category with 98.5 $\pm$ 0.94 \% accuracy compared to OCEC with 88.3 $\pm$ 8.62 \% accuracy.

\subsection{Discovered Relational Categories}

\begin{figure}
 \centering
 \includegraphics[width=0.6\linewidth]{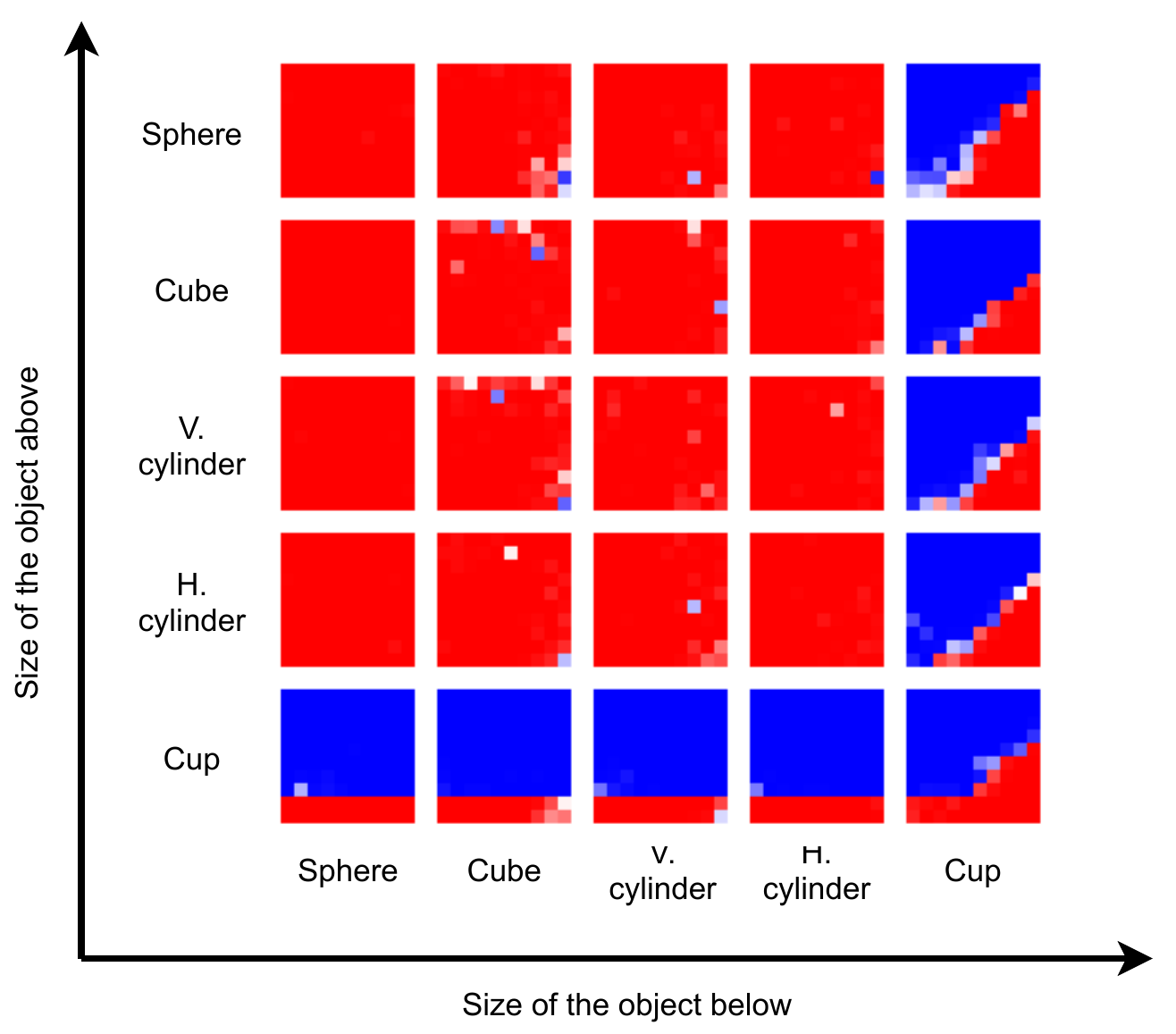}
 \caption{The encoder $f_2$ activations (blue for 0, red for 1) for paired objects. Here, $x$ and $y$ axes of each of the $5\times 5$ plots represent the sizes of the objects below and above, respectively. Each square represents the relation for a given pair. Note that without any direct supervision, the system discovers approximately linear boundaries (e.g., last column) for some object pairs that would help in effect prediction.}
 \label{fig:comparison}
\end{figure}

The stack interaction experience of the robot is used to train the multi-object encoder-decoder network (Figure~\ref{fig:network} bottom), transferring the object categories reported above. The number of binary activated bottleneck neurons is automatically set to 1 using the hyperparameter search described in the Methods section.

The response of the bottleneck neuron, i.e., how this neuron categorizes the input object pairs, is analyzed in Figure~\ref{fig:comparison}. Given different pairs of objects with different sizes, each image in this figure corresponds to a specific object pair, and each pixel provides the response of the bottleneck neuron (0 or 1) for specific object sizes. In our experiments, the effect of stack action depends on object categories and their relative size. For example, if an object is released on top of a larger cup, the released object drops into the cup. If the released object is larger than the cup, it is stacked on top of the cup's walls. The approximately linear boundaries for some object pairs in Figure \ref{fig:comparison} (for example, the last column) show that the bottleneck neuron captured these dynamics and found a symbol that roughly encodes the relative size; the output is 1 when the below cup is larger than the above object. In stacking interactions, the relative size relation only makes sense when the object below is a cup; and our system discovered this relational symbol. Another linear boundary found by the system is in the bottom row. The output of the encoder is 1 when the above object is a cup and below a specific size. We analyze the exploration data to understand why such a boundary emerges. We found out that if the above object is a small cup, the change in the position of the below object is very small.

The learned representations depend on the effect space and the action space of the agent. In our example, after the single-object training stage, the system differentiates different types of objects but does not differentiate different sizes of objects as they are not sufficiently important for the prediction of push actions. Only after it is trained with new data consisting of a new action, namely stacking, does the system start to differentiate between different sizes of cups. The agent only learns richer representations, and therefore better rules, when it has access to a richer action repertoire. This is a desired property of our system as it learns a minimal set of representations needed to predict the outcomes of its actions.

\subsection{Discovered Effect Categories}
\begin{figure}[t]
    \centering
    \begin{subfigure}[b]{0.49\textwidth}
        \centering
        \includegraphics[width=\textwidth]{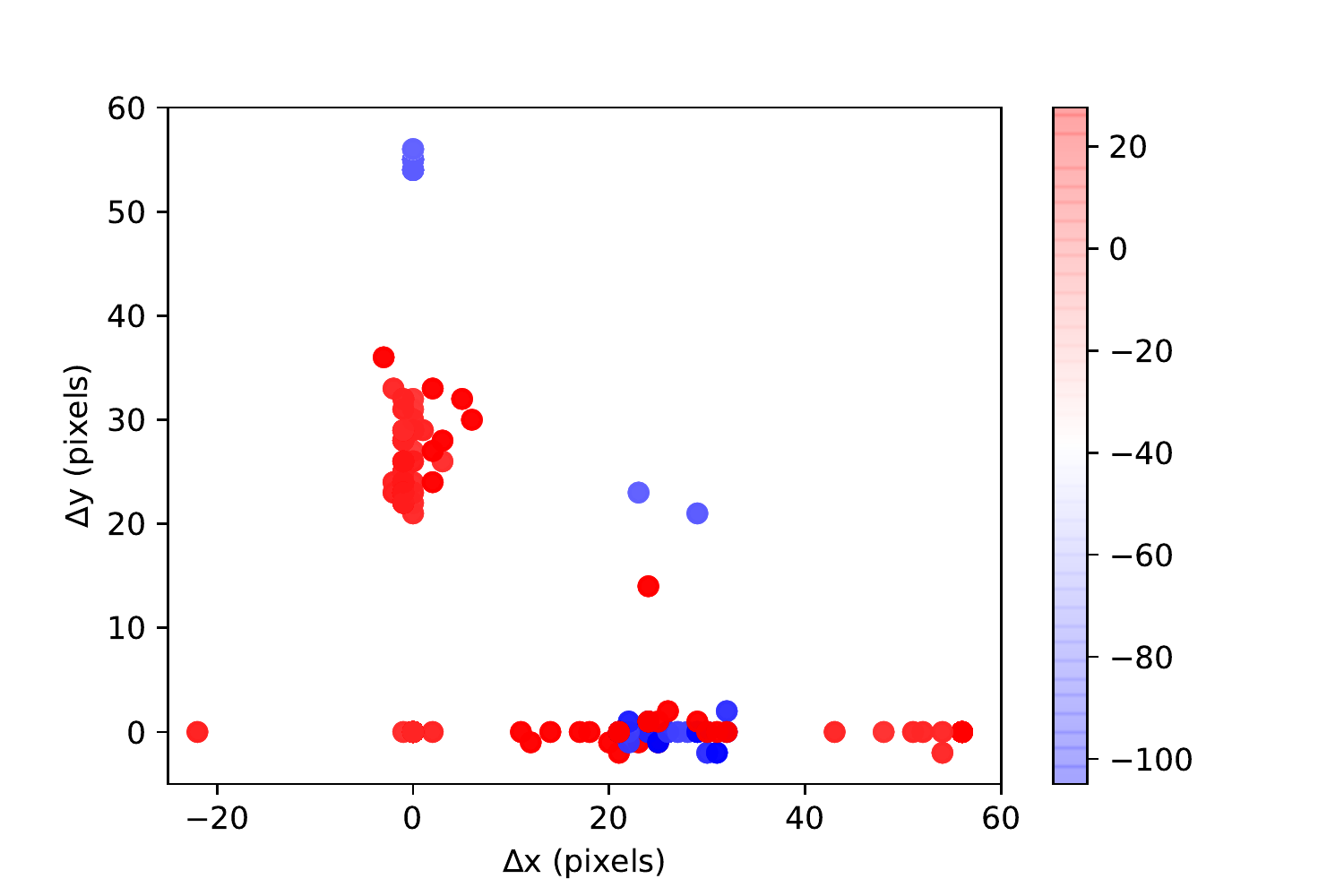}
        \caption{Observed effects, $\mathcal{E}_{\textnormal{single}}$.}
        \label{fig:eff1:obs}
    \end{subfigure}
    \hfill
    \begin{subfigure}[b]{0.49\textwidth}
        \centering
        \includegraphics[width=\textwidth]{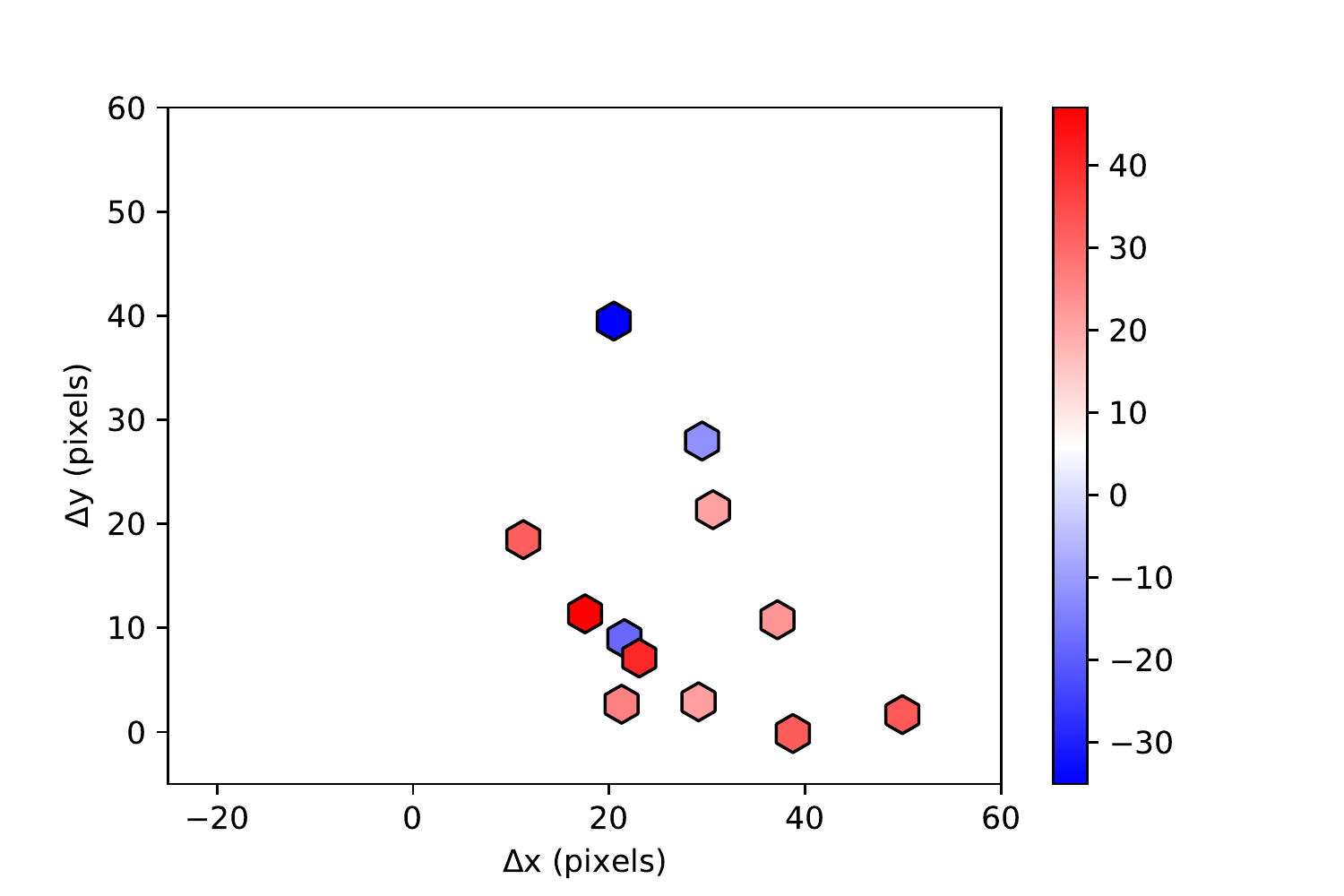}
        \caption{Found effect categories, $\mathcal{C}_{\textnormal{single}}$.}
        \label{fig:eff1:clst}
    \end{subfigure}
    \caption{Effect space for the single object interactions. The low force values are in blue, and the high force values are in red. Note that the found effect categories faithfully represent the effect space without any clustering.}
    \label{fig:eff1}
\end{figure}

\begin{figure}
    \centering
    \begin{subfigure}[b]{0.49\textwidth}
        \centering
        \includegraphics[width=\textwidth]{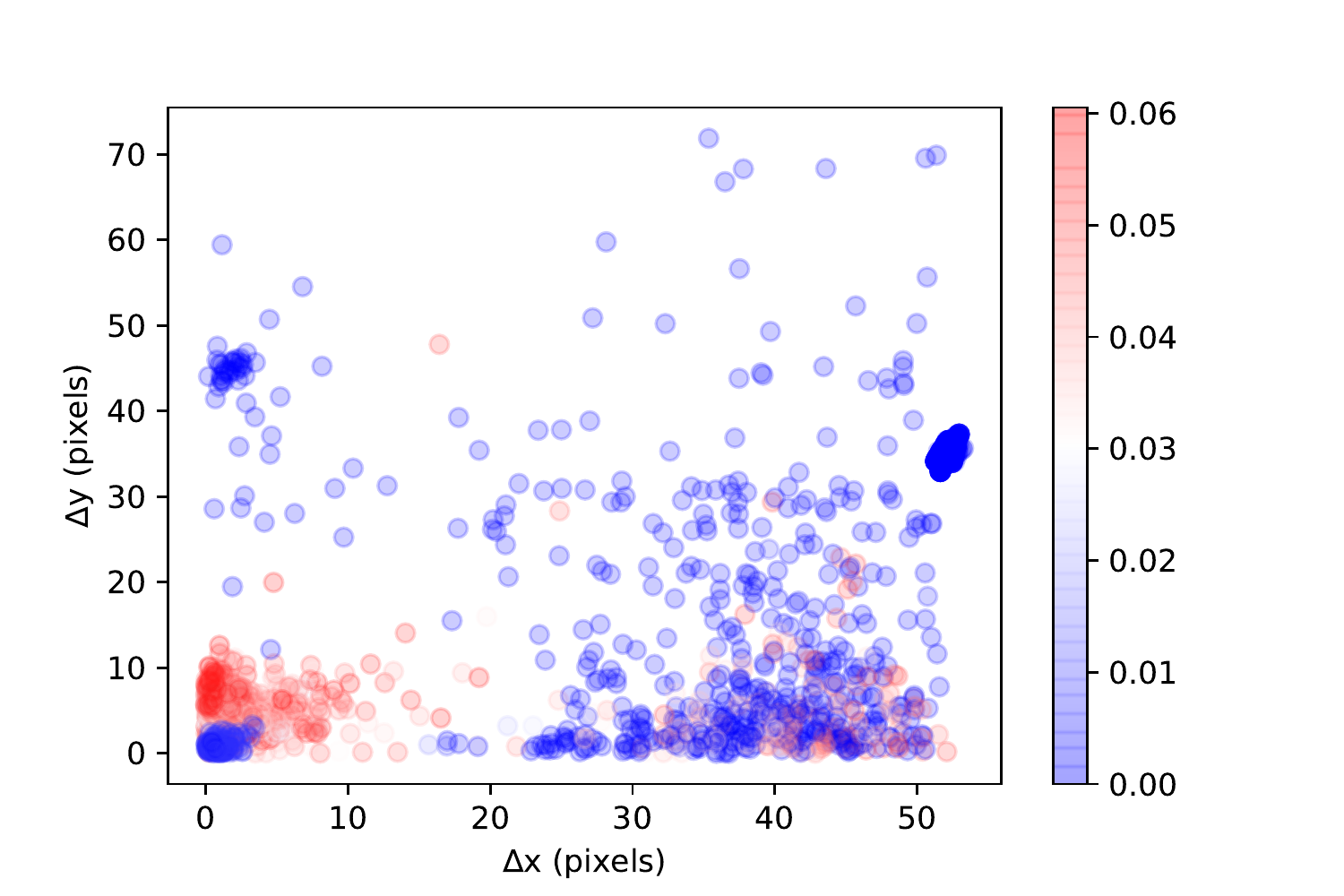}
        \caption{Observed effects, $\mathcal{E}_{\textnormal{paired}}$.}
        \label{fig:eff2:obs}
    \end{subfigure}
    \hfill
    \begin{subfigure}[b]{0.49\textwidth}
        \centering
        \includegraphics[width=\textwidth]{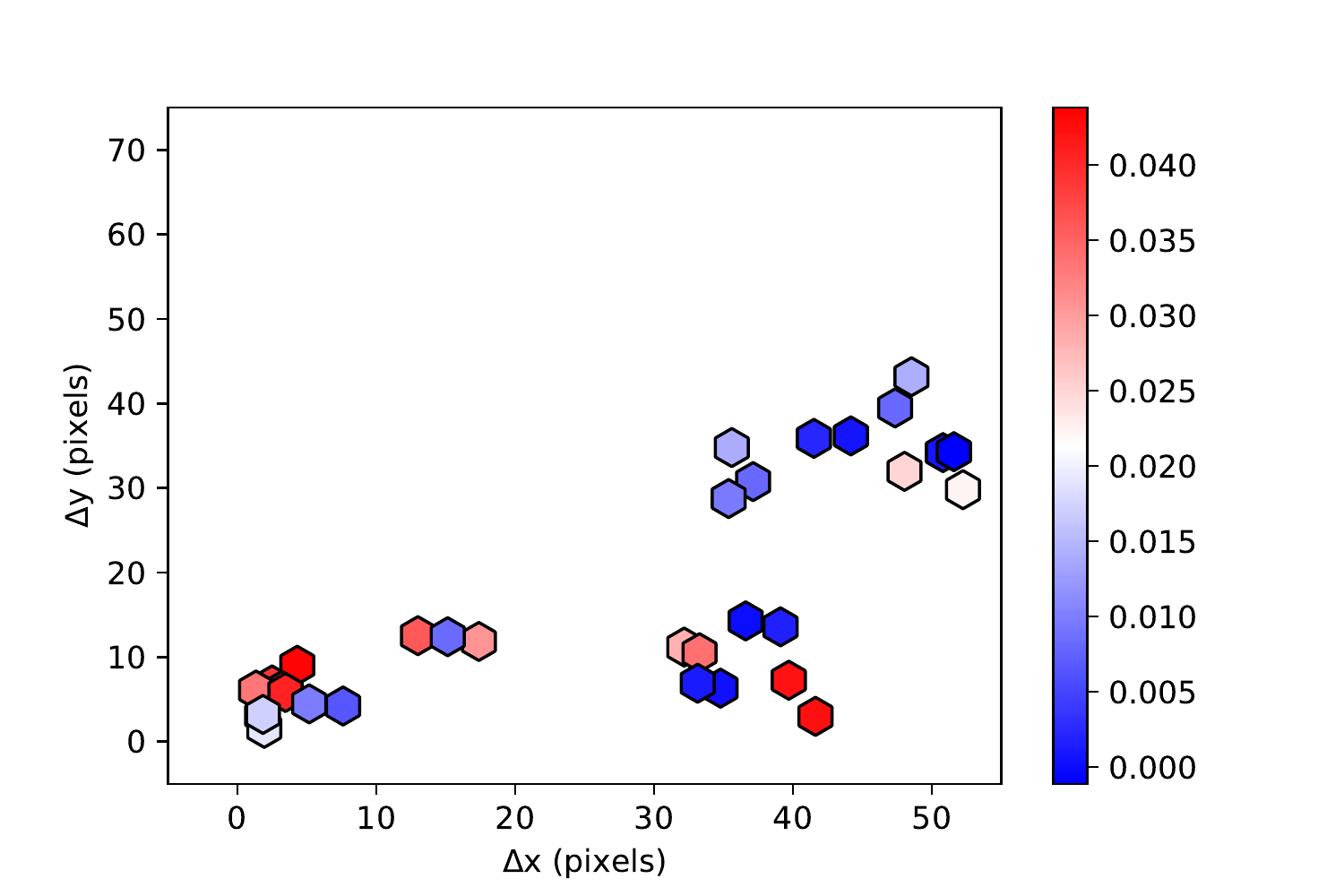}
        \caption{Found effect categories, $\mathcal{C}_{\textnormal{paired}}$.}
        \label{fig:eff2:clst}
    \end{subfigure}
    \caption{Effect space for the paired object interactions. Effects of the below object ($\Delta x_2, \Delta y_2, \Delta d_2$) are omitted as they are almost zero. The low depth values are in blue, and the high depth values are in red.}
    \label{fig:eff2}
\end{figure}

After training, we pass the symbol space $\mathcal{Z}$ together with the action space $\mathcal{A}$ to the decoder to get the effect categories. More specifically:
\begin{align}
    \mathcal{C}_{\textnormal{single}} &= g_1(\mathcal{Z}_{\textnormal{single}}, \mathcal{A}_{\textnormal{single}})\\
    \mathcal{C}_{\textnormal{paired}} &= g_2(\mathcal{Z}_{\textnormal{paired}}, \mathcal{A}_{\textnormal{paired}})\\
\end{align}
Here, $\mathcal{Z}_{\textnormal{paired}}$ is the Cartesian product of the object category space $\{0, 1\}^2$ with the action space $\mathcal{A}_{\textnormal{single}}=\{(0, 0, 1), (0, 1, 0), (1, 0, 0)\}$ resulting in 12 different effect categories for the single object effects. For the paired object effects, the input consists of two single object categories and one relational object category. Therefore, this number is $\{0, 1\}^2 \times \{0, 1\}^2 \times \{0, 1\} \times \mathcal{A}_{\textnormal{paired}} = 32$. Here, $A_{\textnormal{paired}}$ only contains the stack action, therefore $n(\mathcal{A}_{\textnormal{paired}})=1$. These effect categories for the single and the paired interactions are shown in Figures \ref{fig:eff1} and \ref{fig:eff2}, respectively. For visualization purposes, we use colors to represent the third dimension. In Figure \ref{fig:eff1}, the low force values are in blue, and the high force values are in red. Likewise, in Figure \ref{fig:eff2}, the low depth values are in blue, and the high depth values are in red.

\subsection{Learned Rules and PPDDL Operators}
\label{subsec:learned_pddl}

\begin{figure}[t]
    \centering
    \begin{subfigure}[b]{0.49\textwidth}
        \centering
        \includegraphics[width=\textwidth]{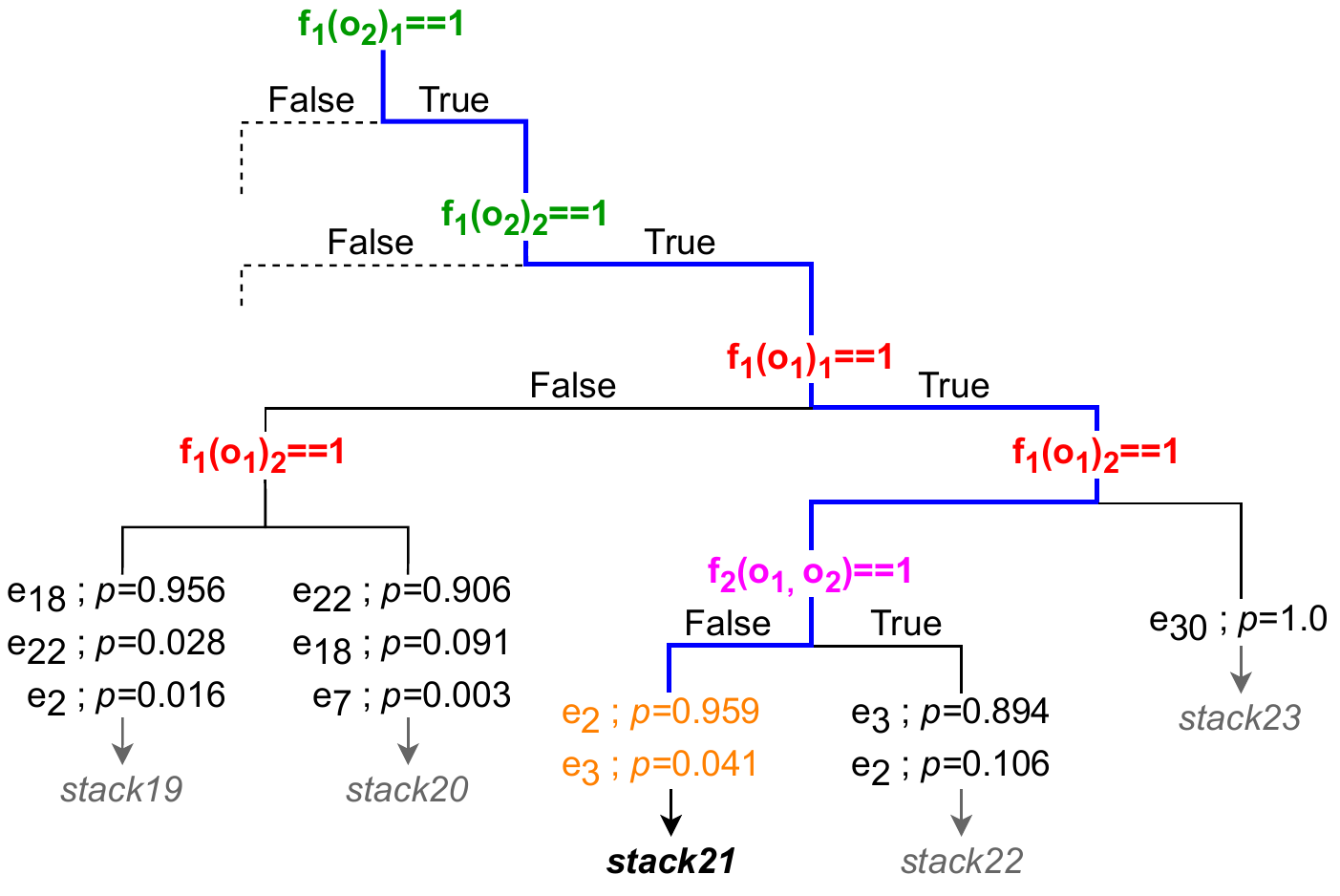}
        \vspace{1em}
        \caption{One path of the decision tree is highlighted.}
        \label{fig:tree}
    \end{subfigure}
    \hfill
    \begin{subfigure}[b]{0.49\textwidth}
        \centering
        \includegraphics[width=\textwidth]{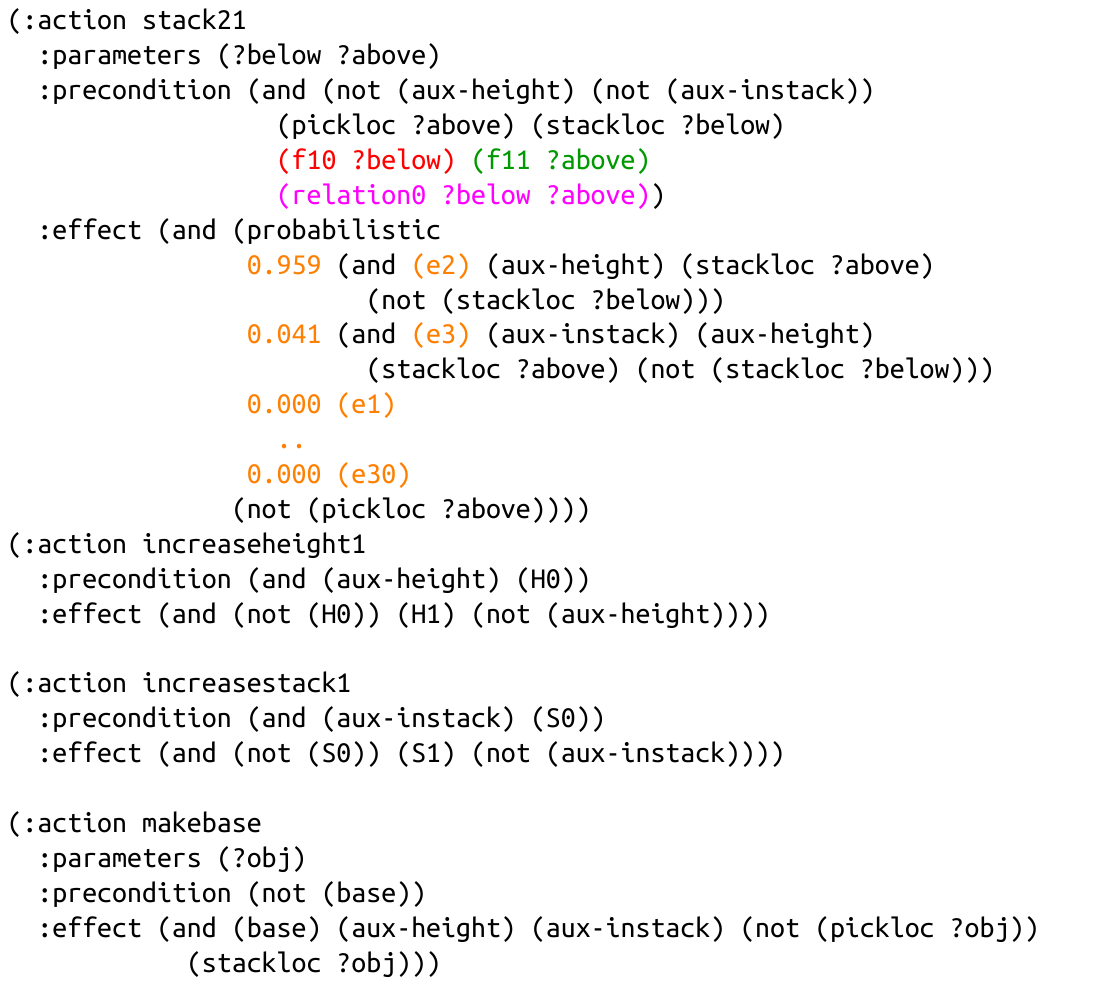}
        \caption{Highlighted path is converted into PPDDL.}
        \label{fig:ppddl}
    \end{subfigure}
    \caption{An example expansion of the decision tree. ($f_1(o_1)_1, f_1(o_1)_2$) represents the category of the object above where ($f_2(o_2)_1, f_2(o_2)_2f$) represents the category of the object below. The relational variable is denoted as $f_2(o_1, o_2)$. On the leaves, each effect $e_i$ is observed with the corresponding probability.}
    \label{fig:ruleconvert}
\end{figure}

The single- and paired-object categories (acquired from the output of the encoder) together with the action vector are used as inputs to the decision tree in order to predict the effect categories (extracted from the output of the decoder). The learned tree is of depth 5, has 24 leaves, and its classification accuracy is 94.8\%. The result of decision tree learning is shown in Figure~\ref{fig:tree}, where only a small number of decision paths out of 24 is explicitly shown because of the space constraints. Decision rules for the highlighted path is $(f_1(o_1)_1, f_1(o_1)_2, f_1(o_2)_1, f_1(o_2)_2, f_2(o_1, o_2))=(1, 0, 1, 1, 0)$. Here, $(f_1(o_1)_1, f_1(o_1)_2)$ represents the category of the object above, $(f_1(o_2)_1, f_1(o_2)_2)$ represents the category of the object below, and $f_2(o_1, o_2)$ is the symbol for the paired-object relation. A natural-language translation of this path is as follows: `If the above object is rollable in all directions (1, 1), and the below object is pushable and insertable (1, 1), and the below object is not larger than the above object, $e_2$ is observed (which is a stacking effect) with 0.959 probability'. PPDDL description corresponding to this decision path of the tree is shown in Figure~\ref{fig:ppddl}.

\begin{figure}[t]
 \centering
 \includegraphics[width=0.99\textwidth]{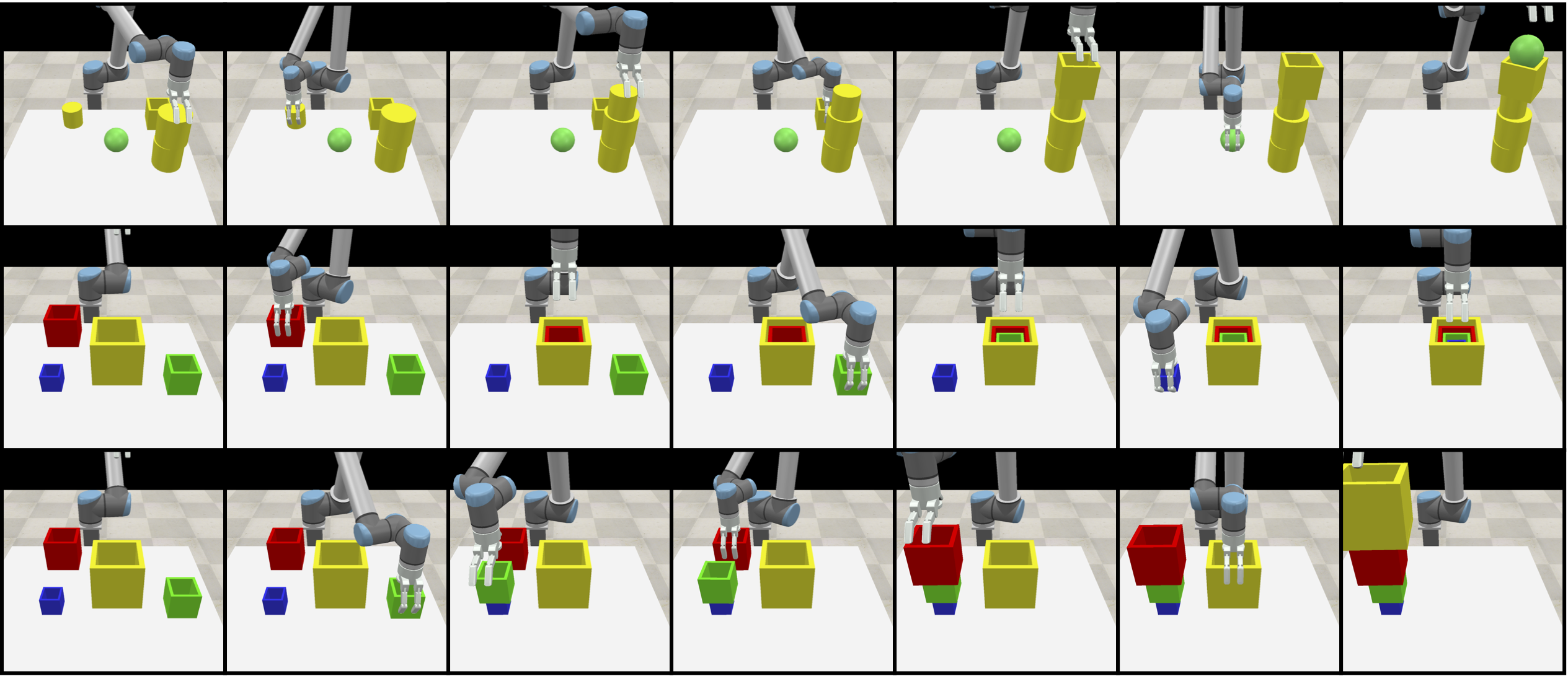}
 \caption{Top row: The objective is to construct a tower of height five using five objects, H5S5. The system assesses the success probability to be 0.07. Middle row: The objective is H1S4, and the system assesses the success probability to be p=0.76. Bottom row: If we change the objective to H4S4, the success probability increases to 0.88.}
 \label{fig:plan}
\end{figure}

For our experiments in the tower building task, we manually introduced some auxiliary predicates, as well as special actions for the domain, to be able to chain multiple actions and count the number of objects in the tower. These are needed to set a goal of constructing a tower with multiple objects which are outside the experience of the robot.

For effects with small $\Delta x_1$ and $\Delta y_2$ the \texttt{aux-instack} predicate is set to true if they satisfy $\Delta d_1>\epsilon$ for some threshold $\epsilon$, and otherwise the \texttt{aux-height} predicate is set to true. For this specific application, these predicates allow us to differentiate stacking and inserting, our effects of interest, from other effects. Actions \texttt{increaseheight1} and \texttt{increasestack1} are treated as addition operators that increase the height of the tower (H) and the number of objects in the stack (S), respectively. There are multiple H and S predicates ranging from H1-H7 and S1-S7, and likewise multiple \texttt{increaseheight} actions. When the \texttt{aux-height} effect is observed, the planner must select the \texttt{increaseheight1} action to continue with the plan. Therefore, when a stack effect is observed, the height of the tower (which is represented by H) increases automatically.

These would not be needed if we were to use the numeric values associated with effect clusters (i.e., functions in PDDL) as then we would be able to set arbitrary goals such as $\Delta d_1 > 0.3$ (`make the height of the tower taller than 30cm'). We went on with the atomic effect representation (i.e., with no associated parameters) as they worked better with the mGPT implementation\footnote{\texttt{https://github.com/bonetblai/mini-gpt}} that we used. In future work, we plan to extend the planner so that we can also utilize numeric values associated with effect clusters.

Lastly, the predicate \texttt{pickloc} is true for objects that are on the table and available for use for the tower construction; \texttt{stackloc} is true for the object that is at the top of the tower. These are shown in Figure~\ref{fig:ppddl}.

\subsection{Performance of Planning}

The PPDDL descriptions that are automatically constructed by the discovered symbols and rules were verified by generating plans given a set of goals, executing these plans in the simulator, and assessing the success of the executed action sequences in achieving these goals. To be concrete, we asked the system to generate plans to create towers of desired heights with a given fixed set of objects. The challenge of the task is to place objects on top of each other in the correct order. With five objects, there are $5!$ plans. For plan generation, the mGPT off-the-shelf probabilistic planner \shortcite{bonet2005mgpt} with FF heuristic \shortcite{hoffmann2001ff} was used.

Since in our experiments, the system is asked to generate plans given a number of objects on the table, we encode the task of, say, ``construct a tower with a height of three (H3) using four objects (S4)'' as H3S4 (Figure~\ref{fig:plan}).

\subsubsection{DeepSym vs. OCEC}

\begin{table}[t]
    \centering
    \begin{tabular}{l|ccc}
        \hline
        & Estimated prob. & Planning success & Execution success \\
        \hline
        DeepSym & 80.4 & 95.0 & 70.0 \\
        OCEC & 12.1 & 25.0 & 15.0\\
        \hline
    \end{tabular}
    \caption{Planning results from random 20 configurations.}
    \label{tab:comparison_plan}
\end{table}

We first compare our system with the alternative OCEC system. We train both systems ten times and select the best-performing models (based on the decision tree accuracy). We initialized 20 random problems and asked the planner to solve the task using two different domain descriptions generated from different methods. We run the probabilistic planner 100 times for each problem and record the number of successes to estimate the success probability of the plans. The results are reported in Table~\ref{tab:comparison_plan}. We report two different metrics: (1) planning success shows whether the system generated a feasible plan or not, and (2) execution success shows whether the execution is successful or not. The latter is concerned with the stochasticity of the environment, not with the feasibility of the plan. We see that the OCEC model performs considerably worse with 25\% planning accuracy than our approach with 95\% planning accuracy. This is mainly caused by the wrong classification of the cup object (see Table \ref{tab:singlecats} in Section \ref{subsec:cat}), which is an essential piece of information in this problem. When single- and paired-object categories are incorrectly classified, the system generates an invalid problem description, which results in infeasible plan outputs. For the same number of symbols, the learned symbols in the OCEC pipeline do not directly depend on the action and the generated effects, while symbols learned with our architecture directly depend on the action and its corresponding effect as they are directly used for effect prediction. This leads to the creation of symbols that are more appropriate for planning.

\subsubsection{DeepSym Performance}
\begin{table}[t]
 \centering
 \begin{tabular}{l|cccc}
 \hline
 Task & H4S4 & H3S4 & H2S4 & H1S4\\
 \hline
 Estimated execution probability & 0.91 & 0.93 & 0.86 & 0.68 \\
 \hline
 Success & 0.88 & 0.56 & 0.68 & 0.32\\
 Planning fail & 0.12 & 0.20 & 0.20 & 0.32 \\
 Execution fail & 0.00 & 0.24 & 0.12 & 0.36 \\
 \hline
 \end{tabular}
 \caption{Planning results from 25 executions for each task. To satisfy the H1S4 objective, the robot needs to insert objects inside each other, which is more challenging compared to the other tower tasks. Thus, the success probability is lower.}
 \label{tab:results_plan}
\end{table}

Now that we have shown the performance gap between the two methods, we want to analyze when our method fails and succeeds. We considered four different goals (towers of heights from 1 to 4) and performed 25 different runs with random initial object configurations for each objective. We configured object types and sizes to have at least one feasible solution. For example, for the H1 objective, we make sure that there are at least three cups that can be physically stacked into each other. The plan execution performance is reported in Table~\ref{tab:results_plan}.
There are three different outcomes: (1) the plan is executed successfully, (2) the planner outputs an erroneous plan due to a recognition error in the encoder, (3) the generated plan is correct, but the plan fails at execution time due to the stochasticity of the environment.

We see that the robot constructs towers with a height of four successfully. As the height of the tower decreases, the robot needs to insert some of the objects inside other cups. The insertion task is harder than the stacking task due to the stochasticity of the environment, which is also reflected in the estimated probabilities in Figure~\ref{fig:ppddl}; even if the below cup is larger than the above sphere, the insertion probability is 0.894. For example, for the challenging objective of creating an H1 tower including all objects, the system estimates the success probability to be 0.68, and therefore the failure probability to be 0.32. Accordingly, 36\% of plans fail at execution time. This shows that the system can partially model the probabilistic nature of the environment. The planning errors are mostly due to the incorrect recognition of the paired-object categories. Example executions are shown in Figure~\ref{fig:plan}.

\subsubsection{Deterministic vs. Probabilistic Planning}
\label{sec:detvsprob}
We also experimented with deterministic planning instead of a probabilistic one. To do so, while converting rules to PDDL, we take the maximum likely effect as the generated effect. For example, if a specific action schema produces effect $e_2$ with a probability of 0.91 and $e_{17}$ with a probability of 0.09, we take the effect with the maximum probability as the generated effect for the action schema. Thus, deterministic planning eliminates the possibility of other effects and, therefore, effectively eliminates possible solutions. When the learned rules faithfully represent the action-effect relations in the environment, we observe no significant difference between probabilistic and deterministic planning in terms of the success of plans. However, when there is significant inaccuracy in the learned representation (e.g., an incorrect comparison between a pair of objects), the probabilistic PDDL description can account for this inaccuracy in the probabilities of effects; the inaccuracies are reflected as the uncertainty of the environment.

\section{Experiments on 8-puzzle}
\label{sec:8puzzle}

\begin{figure}
    \centering
    \includegraphics[width=0.8\textwidth]{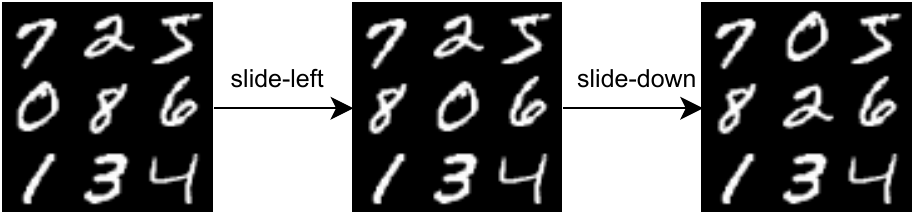}
    \caption{Two steps of the MNIST 8-puzzle. The 0-tile is treated as the empty tile. Each tile consists of a $28\times28$-pixel MNIST digit.}
    \label{fig:mnist8puzzle}
\end{figure}

In this section, we evaluate DeepSym on the MNIST 8-puzzle adapted from \shortciteA{asai2017classical}. In the original 8-puzzle, the aim is to have the tiles in a specific arrangement (considered the goal configuration) by moving tiles into the empty square. In the adapted MNIST 8-puzzle version, tiles do not have symbolic values such as digits but instead contain images of digits, and the 0-tile is treated as the empty tile. 
Given the domain definition, i.e., the knowledge of how the configuration changes in response to slide actions, the 8-puzzle game can be solved with AI planners. However, the problem becomes non-trivial when the states are represented with raw images of the board, and the state transitions are not known. 
In the adapted MNIST 8-puzzle, our system is given the raw image of the board with $(3\times28)\times(3\times 28)=7056$ pixels. Therefore, the state vector is 7056-dimensional. An instance of the MNIST 8-puzzle is shown in Figure \ref{fig:mnist8puzzle}. A system that can solve the puzzle should recognize the following:
\begin{enumerate}
    \item Actions only modify some part of the image (i.e., there are tiles),
    \item There are specific symbolic representations in these tiles (i.e., recognize the image content of the tiles),
    \item The goal is only valid when these tiles are arranged in a specific order (sorted from left to right and top to bottom).
\end{enumerate}

As in our robot experiments, the general pipeline (Figure \ref{fig:pipeline}) consists of four stages: (1) exploration, (2) symbol learning, (3) rule learning, (4) and the translation of rules to PDDL.

\begin{figure}
    \centering
    \hspace{2em}
    \includegraphics[width=0.85\textwidth]{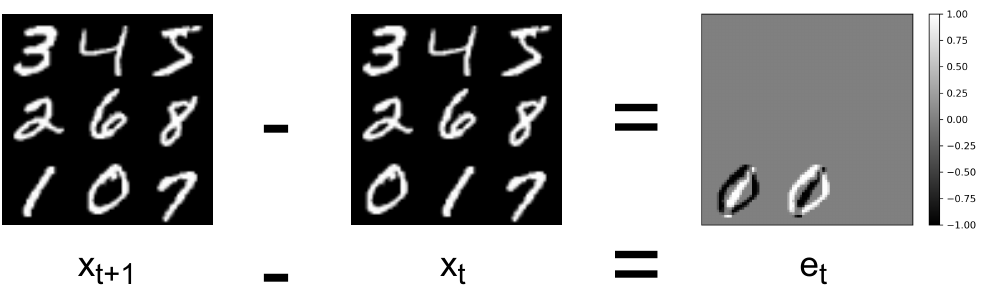}
    \caption{The effect is represented as the difference between two timesteps.}
    \label{fig:8puzzle_effect}
\end{figure}

(1) In the exploration stage, the system initializes a random environment configuration, executes a random action (which is provided to the system), and records a 3-tuple $(\mathbf{x}_t, a_t, \mathbf{e}_t)$ where $\mathbf{x}_t$ is the current state, $a_t$ is the executed action represented as a one-hot vector, and $\mathbf{e}_t$ is the generated effect represented as the pixel difference between the new state $\mathbf{x}_{t+1}$ and the current state $\mathbf{x}_t$ (Figure \ref{fig:8puzzle_effect}). We collect 100,000 such interactions from the environment.
(2) In the symbol learning stage, we train an encoder-decoder network as in Section \ref{subsec:deepnets}, where the encoder $f(\mathbf{x})$ takes the state vector $\mathbf{x}$ as an image of $84\times84$ pixels and outputs a binary vector $\mathbf{z}$, and the decoder $g(\mathbf{z}, a)$ takes the concatenation of $\mathbf{z}$ and the action vector $a$ to produce the effect $\mathbf{e}$ which is also an image of $84\times84$ pixels. Both the encoder and the decoder are convolutional networks. We did not employ any hyperparameter search on the architectures but followed the building principles in DCGAN \shortcite{radford2015unsupervised}. The details of the networks can be found in Appendix \ref{app:network}. We train the model for 100 epochs with MSE loss in Equation \ref{eq:mse}.
(3) After training, we distill the information in the decoder network into rules by training a decision tree using the predictions of the decoder. (4) Lastly, we translate the rules represented by the decision tree into PDDL rules as in Section \ref{subsec:rules}.

\subsection{Learned symbols}
\label{subsec:8puzzle_symbols}

In the MNIST 8-puzzle environment, there is a finite set of possible effects that can be generated in a single action from any environment configuration. If we use the same image for a digit as in \shortciteA{asai2017classical}, then the encoder should represent 3248 different states (digits that are nearby the empty tile) in order for the decoder to produce the correct effect. Since we are using binary activations, $\log_2 3248 \approx 11.67$ units are necessary to avoid losing any information regarding effects. Therefore, we set the number of units to 13 (giving one more as a slack) in this experiment.

\begin{figure}
    \centering
    \includegraphics[width=\textwidth]{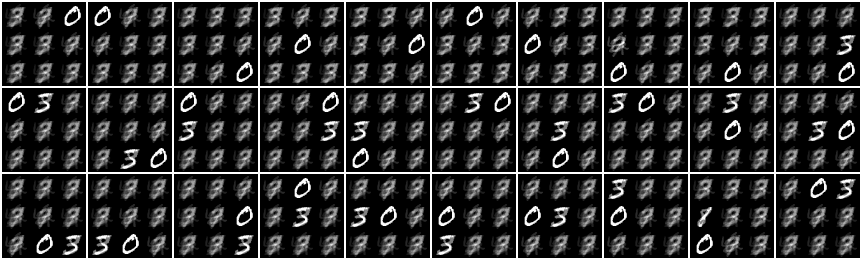}
    \caption{Average states that correspond to the top 30 symbols on MNIST 8-puzzle (sorted by their activation count from left to right and top to bottom)}
    \label{fig:mnist_symbols_3p}
\end{figure}

To understand the symbols that correspond to the low-level subsymbolic representations (i.e., images), we sample 100,000 random states from the environment and get their symbolic representations from the encoder. Then, we take the average of images that correspond to the same symbol. We show the average image that corresponds to the top 30 symbols sorted by their activation counts in Figure \ref{fig:mnist_symbols_3p}. We notice that the first nine symbols correspond to different locations of the empty tile (recall that the digit `0' is considered as the empty tile following \shortcite{asai2017classical}), which accounts for 41.5\% percent of all activations (i.e., in 41.5\% of the time, the encoder only outputs the position of the empty tile). Other symbols correspond to cases where the digit `3' or `5' is near the empty tile.

\begin{figure}[htbp]
    \centering
    \includegraphics[width=0.7\textwidth]{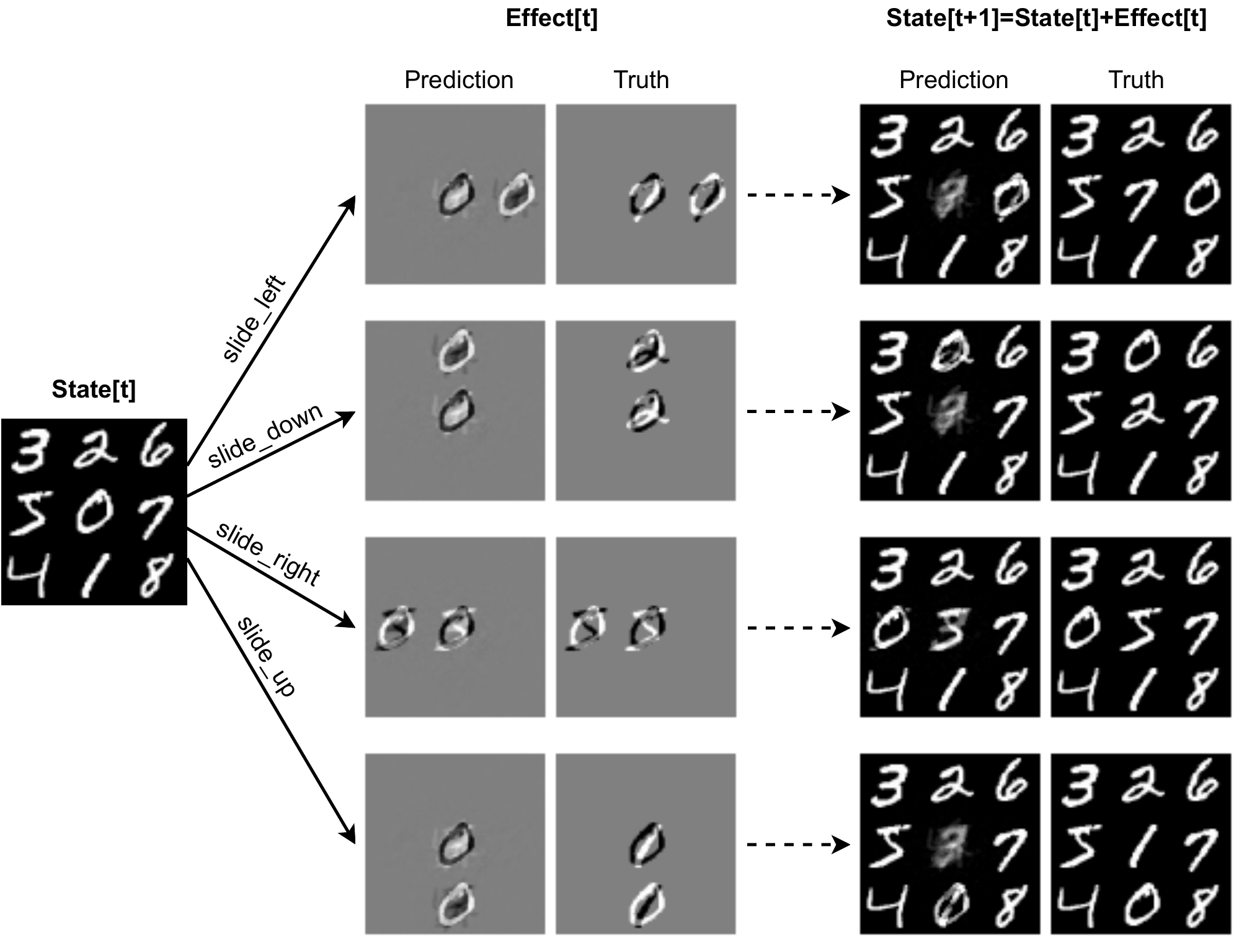}
    \caption{Four different effect predictions are shown together with their ground truths for different actions for a given state. For example, for the `slide-right' action, in the center, `0' is erased and `5' is painted, and on the left, `5' is erased and `0' is painted.}
    \label{fig:mnist_effects_3p}
\end{figure}

In Figure \ref{fig:mnist_effects_3p}, some predicted effects are visualized for a given state and actions together with the ground truth effects. We see that the decoder successfully models the slide of the digit `0'. When we combine the previous state with the predicted effect, we can have an estimate of the next state which is shown in the right column in Figure \ref{fig:mnist_effects_3p}.

\subsection{Learned Rules}

To train a decision tree for the rule extraction, we collect the set of training examples as follows. Given the current state $\mathbf{x}_t$, the encoder generates the corresponding symbol $\mathbf{z}_t = f(\mathbf{x}_t)$ which is then used as input to the decoder together with the one-hot action vector $a_t$ to predict the effect: $\bar{\mathbf{e}}_t = g(\mathbf{z}_t, a_t)$. Then, we predict the next state $\mathbf{x}_{t+1}$ by summing the predicted effect $\bar{\mathbf{e}}_t$ with the current state ($\bar{\mathbf{x}}_{t+1} = \mathbf{x}_t + \bar{\mathbf{e}}_t$) as in Figure \ref{fig:mnist_effects_3p}. Lastly, we use the encoder to generate the symbol $\bar{\mathbf{z}}_{t+1}$ that corresponds to $\bar{\mathbf{x}}_{t+1}$: $\bar{\mathbf{z}}_{t+1} = f(\bar{\mathbf{x}}_{t+1})\label{eq:z_n}$. The decision tree is trained with $\{[\mathbf{z}_t; a_t], \bar{\mathbf{z}}_{t+1}\}$ input-output pairs. This is even more generic than robot experiments where we trained the tree with $\{[\mathbf{z}_t; a_t], c_t\}$ pairs ($c_t$ is the effect category predicted by the decoder) since it allows us to express the goal using the image modality. In both cases, the fundamental idea is the same: training a decision tree with symbolic input-output pairs in order to learn probabilistic rules.

As the last step, we convert the decision paths of the tree into probabilistic PDDL rules as in Section \ref{subsec:learned_pddl}. As an example, a translated rule from a decision path is as follows,
\begin{verbatim}
(:action slide_left5
    :precondition (and (not (z9)) (not (z5)) (z3))
    :effect (probabilistic
             0.16667 (and (z0) (z1) (not (z2)) (not (z3)) (z4)
                          (z5) (not (z6)) (not (z7)) (z8) (z9)
                          (z10) (not (z11)) (z12))
             0.00758 (and (z0) (z1) (not (z2)) (not (z3)) (z4)
                          (z5) (not (z6)) (not (z7)) (z8) (z9)
                          (z10) (z11) (z12))
             0.68939 (and (z0) (z1) (not (z2)) (z3) (z4) (z5)
                          (not (z6)) (not (z7)) (z8) (z9) (z10)
                          (not (z11)) (z12))
             0.13636 (and (z0) (z1) (not (z2)) (z3) (z4) (z5)
                          (not (z6)) (not (z7)) (z8) (z9) (z10)
                          (z11) (z12))))
\end{verbatim}
where predicates \texttt{(z0)} \dots \texttt{z(12)} correspond to activations of each unit in $z$. There are no auxiliary predicates as there are in the tabletop environment; the PDDL file only consists of such translated rules given above.

\subsection{Planning Examples}
\label{subsec:planning}
\begin{figure}[htbp]
    \centering
    \includegraphics[width=\textwidth]{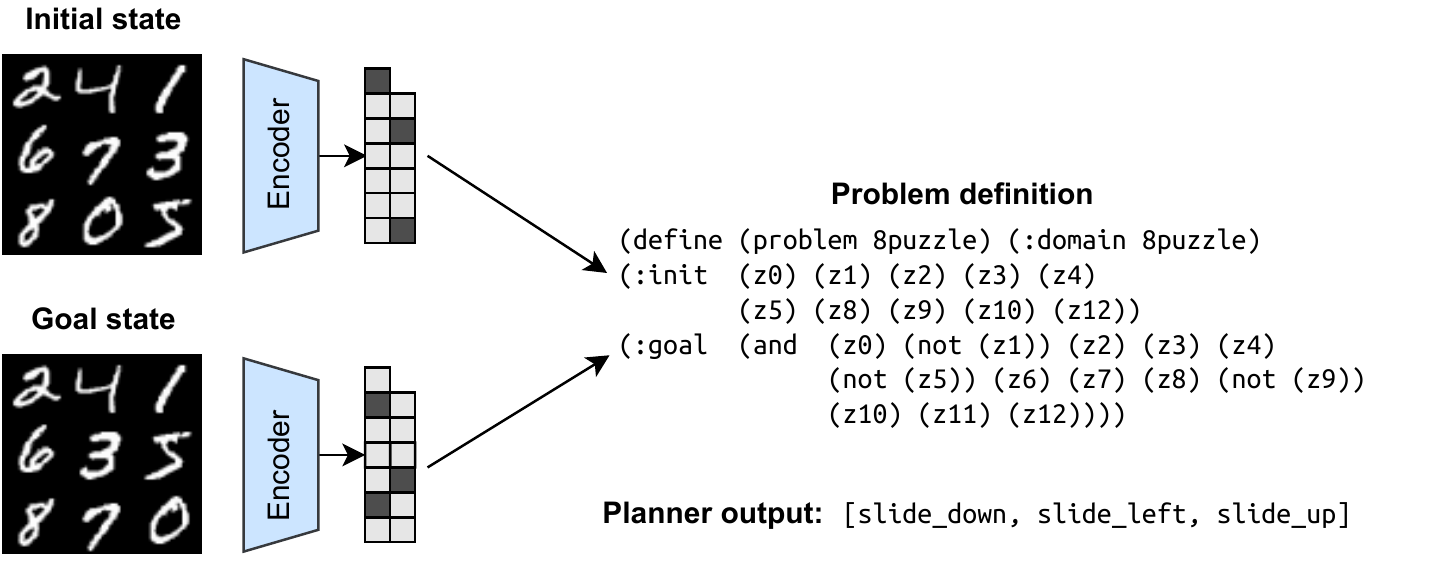}
    \caption{The generated plan for the goal state from a random initial state.}
    \label{fig:8puzzle_plan}
\end{figure}

Using the generated PPDDL description, our system is requested to output a plan for the goal state from a random initial state. For this, the problem definition (where the current state and the goal state are indicated) is created in PPDDL using the activations of the encoder (see Figure \ref{fig:8puzzle_plan}).

As shown in the figure, our system was able to find the correct action sequences in order to reach the given goal configuration. Note that we observed that the output plan only slides the tiles so as to move the empty tile into the correct position. This is a consequence of the system because the encoded activations do not represent the global state but a local state: the position of the empty tile and its neighbors. One can extend the locality by incorporating multiple timestep effects of actions in a similar approach with \shortcite{xu2021deep}. This can be an advantage, or disadvantage, depending on the context and the problem, which will be discussed in Section \ref{sec:discussion}.

\begin{figure}[htbp]
    \centering
    \includegraphics[width=0.8\textwidth]{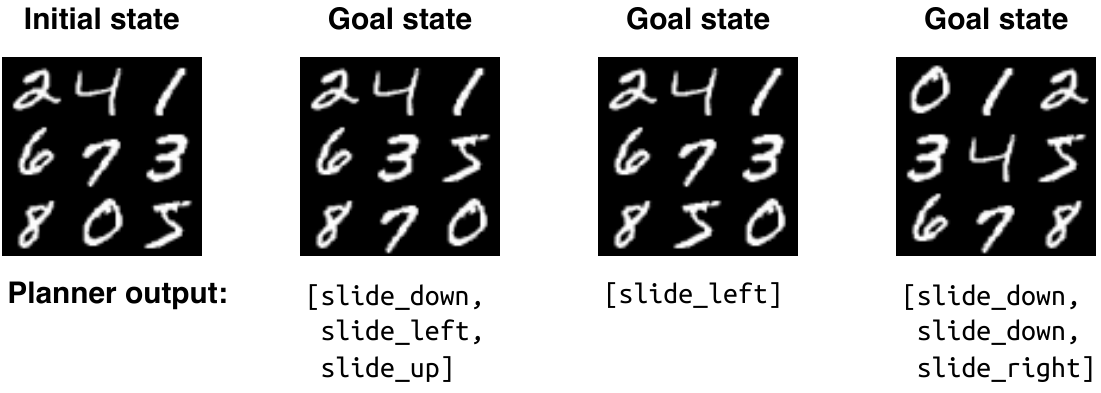}
    \caption{Three different goal positions and their respective planner outputs.}
    \label{fig:twogoals}
\end{figure}

As the current formulation cannot capture the global state, we experimented with local state representations. For example, in Figure \ref{fig:twogoals}, we set two different arbitrary goals that are one step and three steps away from the initial state (the first and the second goal in Figure \ref{fig:twogoals}). The planner outputs the correct plan since it can capture the nearby tile information. However, when asked for the third goal in Figure \ref{fig:twogoals}, the generated plan only moves the empty tile to the correct position while disregarding other tiles.

We generated 100 random goal states that are $n$-step away from the corresponding initial state and reported the planning results to quantitatively assess the performance of the method. We also add the results for executing random actions to assess the performance increment. We report the percentage of plans that successfully move the empty tile to the correct position in Table \ref{tab:8tile_plan_success}. From the results, we see that DeepSym can successfully move the empty tile into the correct position for different plan lengths.
\begin{table}[htbp]
    \centering
    \begin{tabular}{lcccc}
         & 1-step & 2-step & 3-step & 4-step \\
        \hline
        Random plans & $24.4 \pm 4.2$ & $9.8 \pm 2.3$ & $11.4 \pm 2.1$ & $10.6 \pm 4.0$\\
        DeepSym & $92.6 \pm 5.8$ & $88.0 \pm 8.6$ & $88.8 \pm 7.4$ & $89.0 \pm 7.9$\\
        \hline
    \end{tabular}
    \caption{The average percentage of successful plans that move the empty tile for different plan lengths over five runs.}
    \label{tab:8tile_plan_success}
\end{table}

\subsection{Scaling-up to 15-puzzle}
\label{subsec:scaleup}

This section aims to analyze the performance of the system when we scale up the dimensionality of the environment. Examples in the previous section suggest that our system can correctly identify the empty tile, learn the transition based on the empty tile, and make plans to move the empty tile into different positions. We would like to test whether this is the case for larger environments. Therefore, we scale up the 8-puzzle in two different ways: (1) 8-puzzle with replacement (will be denoted as w/r) and (2) 15-puzzle with replacement. Each digit except `0' (the empty tile) may appear more than once in these versions. We train DeepSym with 14 units for 8-puzzle w/r, and with 15 units for 15-puzzle w/r (see Appendix \ref{app:num_states} for details).

\begin{figure}
    \centering
    \begin{subfigure}[b]{0.3\textwidth}
        \centering
        \includegraphics[width=\textwidth]{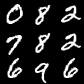}
        \caption{8-puzzle w/replacement.}
        \label{subfig:3x3}
    \end{subfigure}
    \hspace{2em}
    \begin{subfigure}[b]{0.3\textwidth}
        \centering
        \includegraphics[width=\textwidth]{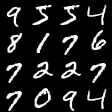}
        \caption{15-puzzle w/replacement.}
        \label{subfig:4x4}
    \end{subfigure}
    \caption{In these environments, each digit except `0' may appear more than once.}
\end{figure}

\begin{figure}
    \centering
    \includegraphics[width=0.8\textwidth]{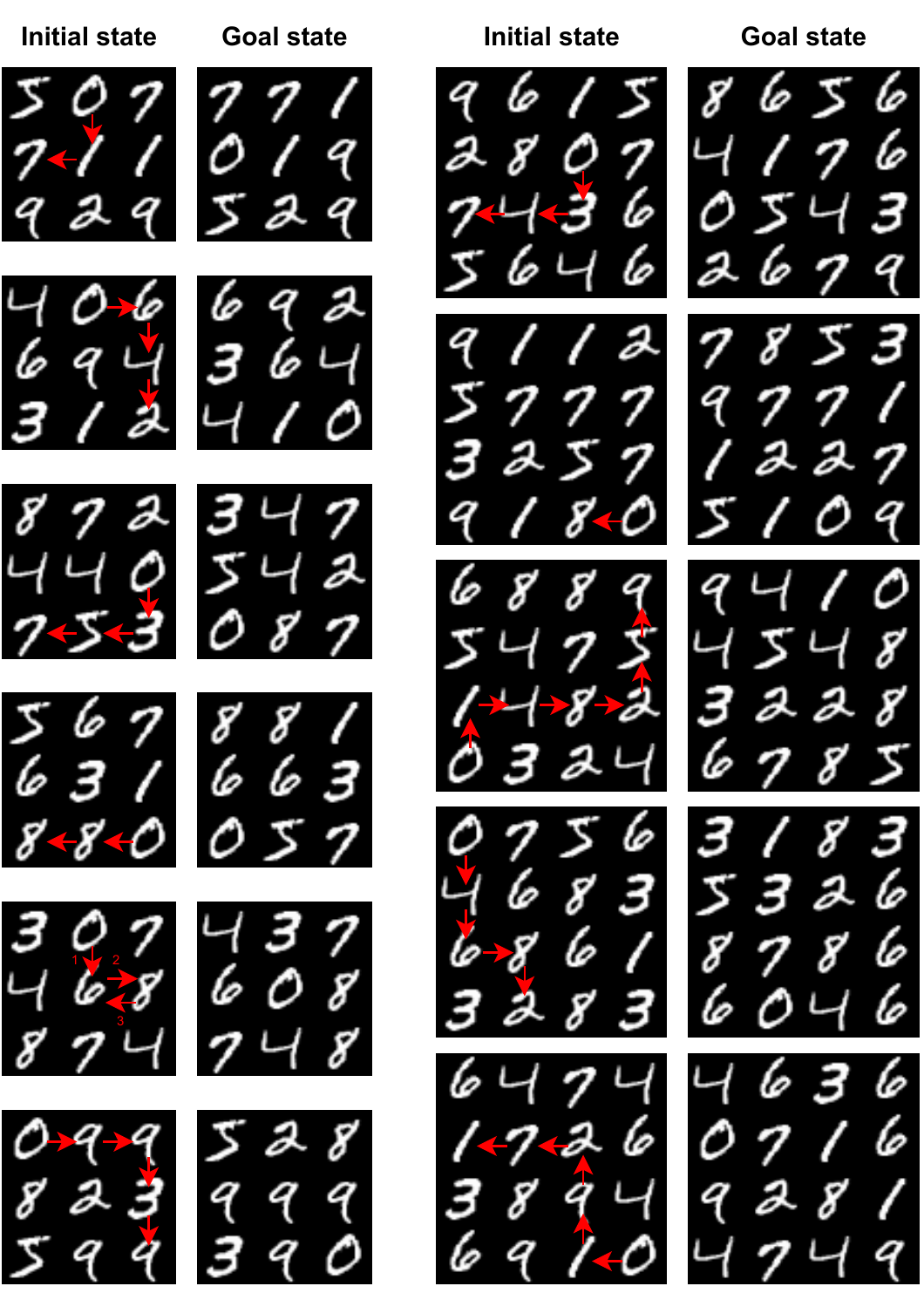}
    \caption{Planning results for 8-puzzle w/r and 15-puzzle w/r. The arrow denotes the movement of the empty tile at each step.}
    \label{fig:15puzzle_plan}
\end{figure}

The low-level state-space and effect-space are $112\times112=12544$ dimensional for 15-puzzle w/r while it stays the same for 8-puzzle w/r. We used the same convolutional architecture with different paddings to ensure the same output size. The most frequently activated symbols are shown in Figures \ref{fig:mnist_symbols_3r} and \ref{fig:mnist_symbols_4r}. We give the planning results for these environments in Figure \ref{fig:15puzzle_plan}. We see that the system moves the empty tile (by sliding other tiles) to the correct position but disregards other tiles.

\subsection{Comparison with Autoencoder}
\label{subsec:comp_ae}

This section aims to compare DeepSym with an autoencoder baseline. We train an autoencoder (as in \shortciteR{asai2017classical}) in these three MNIST $n$-puzzle environments with the same architecture and the same number of bottleneck units as in DeepSym. Given the bottleneck size, it would be impossible for the autoencoder to encode all state-space. The most frequently activated symbols for 8-puzzle, 8-puzzle w/r, and 15-puzzle w/r are shown in Figures \ref{fig:sae_3p}, \ref{fig:sae_3r}, and \ref{fig:sae_4r}, respectively. We train a decision tree for rule learning using the encoder activations to compare the planning performance. Namely, the decision tree is trained with $(f(\mathbf{x}_t), f(\mathbf{x}_{t+1}))$ input-output pairs where $f$ is the encoder network. After the training, we extracted rules from all paths of the tree and constructed a PPDDL description. The planner failed to produce any plan output for random initial and goal states. This is expected since all the state-space cannot be encoded, and therefore, some states are not represented correctly in the PPDDL description. One would need to increase the bottleneck size in order to convert all the state space into PPDDL descriptions. In \shortciteA{asai2017classical}, the bottleneck size is set to 25 units (instead of 13 in our experiments) to cover the state space.

\section{Discussion}\label{sec:discussion}

A plan corresponds to a sequence of actions to move from an initial state to a goal state. One must chain the effects of actions to predict a future state. Thus, the effects of actions should be known to generate a plan. Therefore, the capability of knowing preconditions of actions and predicting the effects of actions is a requirement for generating a successful plan \shortcite{Konidaris2014}.

The main difference between DeepSym and approaches that focus on compressing the state representation (e.g., with an autoencoder, \shortciteR{asai2017classical,asai2022classical}, or with a world model, \shortciteR{hafner2020mastering}) is that the learned representations in DeepSym are only due to actions and effects of the agent \shortcite{Taniguchi2019}. Learning symbols based on the capabilities of the agent allows one to filter-out details of the environment not related to the agent. On the other hand, the approach of compressing the state representation brings its own advantages. One can use a large dataset of states to pre-train an unsupervised model to learn a compact model of the environment, and then use the learned model to train a supervised model for planning or policy learning.

Finding action-independent discrete representations is non-trivial in a large state-space even for the toy examples given in Section \ref{sec:8puzzle}. In our robot experiments, the autoencoder with discrete units \shortcite{asai2017classical} was shown not to generate a useful representation with a low bottleneck dimension. On the other hand, DeepSym can learn useful and compact representations for planning as it considers actions and effects. For environments that are more realistic for lifelong learning, such as Minecraft \shortcite{johnson2016malmo}, the raw state-space is virtually infinite, making it difficult to find a minimal set of meaningful discrete representations without taking actions and action effects into account. On the other hand, action- and effect-based learning allows for an efficient representation of the state space by filtering out the aspects of the environment not relevant to the actions of the agent. For example, in the 8-puzzle environment, the encoder disregards the tiles not near the empty tile since the generated effect does not depend on them. The learned representation allows for generating plans to move the empty tile to different positions. DeepSym learns the minimal set of representations that are needed for the effect prediction of action. Therefore, our system learns action-centric representations, i.e., representations that involve the empty tile. State- and action-based methods are two different (possibly complementary) approaches with advantages and disadvantages. For example, if the problem domain is small, or there exists a large-scale pre-trained model of the environment, encoding all the state space will allow one to solve any encountered problem. However, this approach might be infeasible for larger domains. On the other hand, action-based encoding learns the minimal set of symbols to predict effects at the cost of missing possibly global task requirements (e.g., a specific arrangement in 8-puzzle).

It is theoretically possible to learn a simpler feature-based representation that will be more computationally efficient when compared with deep networks when state, action, and observed effects are all known \shortcite{Ugur-2015-ICRA,konidaris2018skills}. However, this approach would need manual feature extraction for newly encountered domains, while a differentiable network that can be automatically tuned offers a more uniform and extendible approach.

One thing we observed is that with the narrow bottleneck size (i.e., 13 units for 3248 configurations), the encoder does not represent all the neighbor configurations that are needed for effect prediction. However, when we increase the bottleneck size, the encoder indeed learns all the necessary configurations. Even if the system successfully encodes all the local states, it would still need to symbolically encode the global state to solve the task globally. One approach to encoding the global state might be extending the locality by considering the effects of multiple timesteps \shortcite{xu2021deep} or partitioning the state into several chunks and representing the change in these chunks separately.

\section{Conclusion}
\label{sec:conclusion}
In this work, we introduced a method that discovers effect- and action-guided object categories, encodes them as discrete symbols, and learns rules that predict action effects. It sustains a general cognitive development progression where symbols are formed, rules are learned, planning is achieved, and verified in execution. Our system contributes to the state-of-the-art by showing the following desirable properties which have not been achieved/shown simultaneously elsewhere:

\begin{itemize}

 \item We proposed a generic, single pipeline neural solution for mapping raw sensorimotor experience into the symbolic domain.
 
 \item The proposed network allows progressive learning of increasingly complex abstractions, exploiting previously-learned abstractions as inputs.
 
 \item It is gradient-friendly, so it can be incorporated into any gradient-based machine learning system for more complex processing.

 \item When compared with the continuous bottleneck layer version of our system, i.e., OCEC, our system performs better in effect category formation leading to more successful action planning. This suggests that instead of post-training clustering of the continuous unit outputs, employing discrete units from the beginning is beneficial.
 
\end{itemize}

In future work, we plan to scale up the system by augmenting the perceptual capabilities and the action repertoire of the robot. The ad-hoc perceptual system for determining action effects can be replaced by a state-of-the-art computer vision system. Beyond paired-object relations, graph neural networks can be employed to construct relations between varying numbers of objects. Applying the principles of learning and abstraction of this work to less-constrained scenarios will constitute a major step towards AI-enabled, general-purpose robots.

\acks{This research was supported by TUBITAK (The Scientific and Technological Research Council of Turkey) ARDEB 1001 program (project number: 120E274) and by the BAGEP Award of the Science Academy. Additional support is provided by the International Joint Research Promotion Program, Osaka University under the project ``Developmentally and biologically realistic modeling of perspective invariant action understanding'' and the Japan Society for the Promotion of Science, Grant-in-Aid for Scientific Research -- project with number 22H03670. The numerical calculations reported in this paper were partially performed at TUBITAK ULAKBIM, High Performance and Grid Computing Center (TRUBA). We would like to thank Aysu Sayin and Serkan Bugur for their initial help with the experiments. We thank the reviewers of this paper for their detailed comments.}

\appendix

\section{Network Architecture and Hyperparameters}
\label{app:network}
\subsection{Tabletop environment}
The network architectures of encoders $f_1$ and $f_2$ are shown in Tables \ref{tab:encoder1} and \ref{tab:encoder2}, respectively. Each convolution is followed by a batch normalization layer and ReLU activation after the normalization. The network architectures of decoders $g_1$ and $g_2$ are shown in Tables \ref{tab:decoder1} and \ref{tab:decoder2}, respectively. Decoders consist of fully connected (FC) layers with no batch normalization.

\begin{table}[htbp]
\centering
    \begin{subtable}[h]{0.49\textwidth}
        \centering
        \begin{tabular}{|c|c|c|c|c|}
            \hline
            Layer & In ch. & Out ch. & Stride & Pad \\
            \hline
            \hline
            Conv3x3 & 1 & 32 & 1 & 1 \\
            Conv3x3 & 32 & 32 & 2 & 1 \\
            \hline
            Conv3x3 & 32 & 64 & 1 & 1 \\
            Conv3x3 & 64 & 64 & 2 & 1 \\
            \hline
            Conv3x3 & 64 & 128 & 1 & 1 \\
            Conv3x3 & 128 & 128 & 2 & 1 \\
            \hline
            Conv3x3 & 128 & 256 & 1 & 1 \\
            Conv3x3 & 256 & 256 & 2 & 1 \\
            \hline
            \multicolumn{5}{|c|}{Global average pooling over channels}\\
            \hline
            FC & 256 & 2 & - & - \\
            \hline
            \multicolumn{5}{|c|}{Gumbel-sigmoid}\\
            \hline
            \hline
            \multicolumn{5}{|c|}{Number of parameters: 1,174,114}\\
            \hline
        \end{tabular}
        \caption{Encoder $f_1$.}
        \label{tab:encoder1}
    \end{subtable}
    \hfill
    \begin{subtable}[h]{0.49\textwidth}
        \centering
        \begin{tabular}{|c|c|c|c|c|}
            \hline
            Layer & In ch. & Out ch. & Stride & Pad \\
            \hline
            \hline
            Conv3x3 & 2 & 32 & 1 & 1 \\
            Conv3x3 & 32 & 32 & 2 & 1 \\
            \hline
            Conv3x3 & 32 & 64 & 1 & 1 \\
            Conv3x3 & 64 & 64 & 2 & 1 \\
            \hline
            Conv3x3 & 64 & 128 & 1 & 1 \\
            Conv3x3 & 128 & 128 & 2 & 1 \\
            \hline
            Conv3x3 & 128 & 256 & 1 & 1 \\
            Conv3x3 & 256 & 256 & 2 & 1 \\
            \hline
            \multicolumn{5}{|c|}{Global average pooling over channels}\\
            \hline
            FC & 256 & 1 & - & - \\
            \hline
            \multicolumn{5}{|c|}{Gumbel-sigmoid}\\
            \hline
            \hline
            \multicolumn{5}{|c|}{Number of parameters: 1,174,145}\\
            \hline
        \end{tabular}
        \caption{Encoder $f_2$.}
        \label{tab:encoder2}
    \end{subtable}
\end{table}

\begin{table}[htbp]
    \centering
    \begin{subtable}[h]{0.48\textwidth}
        \centering
        \begin{tabular}{|c|c|c|}
            \hline
            Layer &  Input units & Output units \\
            \hline
            \hline
            FC+ReLU & 5 & 128 \\
            \hline
            FC+ReLU & 128 & 128 \\
            \hline
            FC & 128 & 3\\
            \hline
            \hline
            \multicolumn{3}{|c|}{Number of parameters: 17,667}\\
            \hline
        \end{tabular}
        \caption{Decoder $g_1$.}
        \label{tab:decoder1}
    \end{subtable}
    \hfill
    \begin{subtable}[h]{0.48\textwidth}
        \centering
        \begin{tabular}{|c|c|c|}
            \hline
            Layer &  Input units & Output units \\
            \hline
            \hline
            FC+ReLU & 5 & 128 \\
            \hline
            FC+ReLU & 128 & 128 \\
            \hline
            FC & 128 & 6\\
            \hline
            \hline
            \multicolumn{3}{|c|}{Number of parameters: 18,054}\\
            \hline
        \end{tabular}
        \caption{Decoder $g_2$.}
        \label{tab:decoder2}
    \end{subtable}
\end{table}

Adam optimizer \shortcite{kingma2014adam} with AMSGrad \shortcite{reddi2019convergence} is used. The learning rate is set to 0.00005 with a 128 batch size. Each model is trained for 300 epochs, and we select the best model based on the mean square error.

While finding the number of hidden units, we take five runs and record the MSE. We increase the number of units if the one-sided Welch's t-test rejects the null hypothesis $\mathcal{H}_0:$ ``Two numbers result in the same MSE'' in favor of $\mathcal{H}_1:$ ``Increased number results in lower MSE''.

We realize that this requires multiple runs and, in fact, is quite inefficient. It would be better if we had a well-defined metric such as the Bayesian Information Criterion (BIC). We did not use BIC since it does not lead to plausible results with deep neural networks that have large parameter sizes.

\begin{figure}
    \centering
    \begin{subfigure}[b]{0.49\textwidth}
        \centering
        \includegraphics[width=\textwidth]{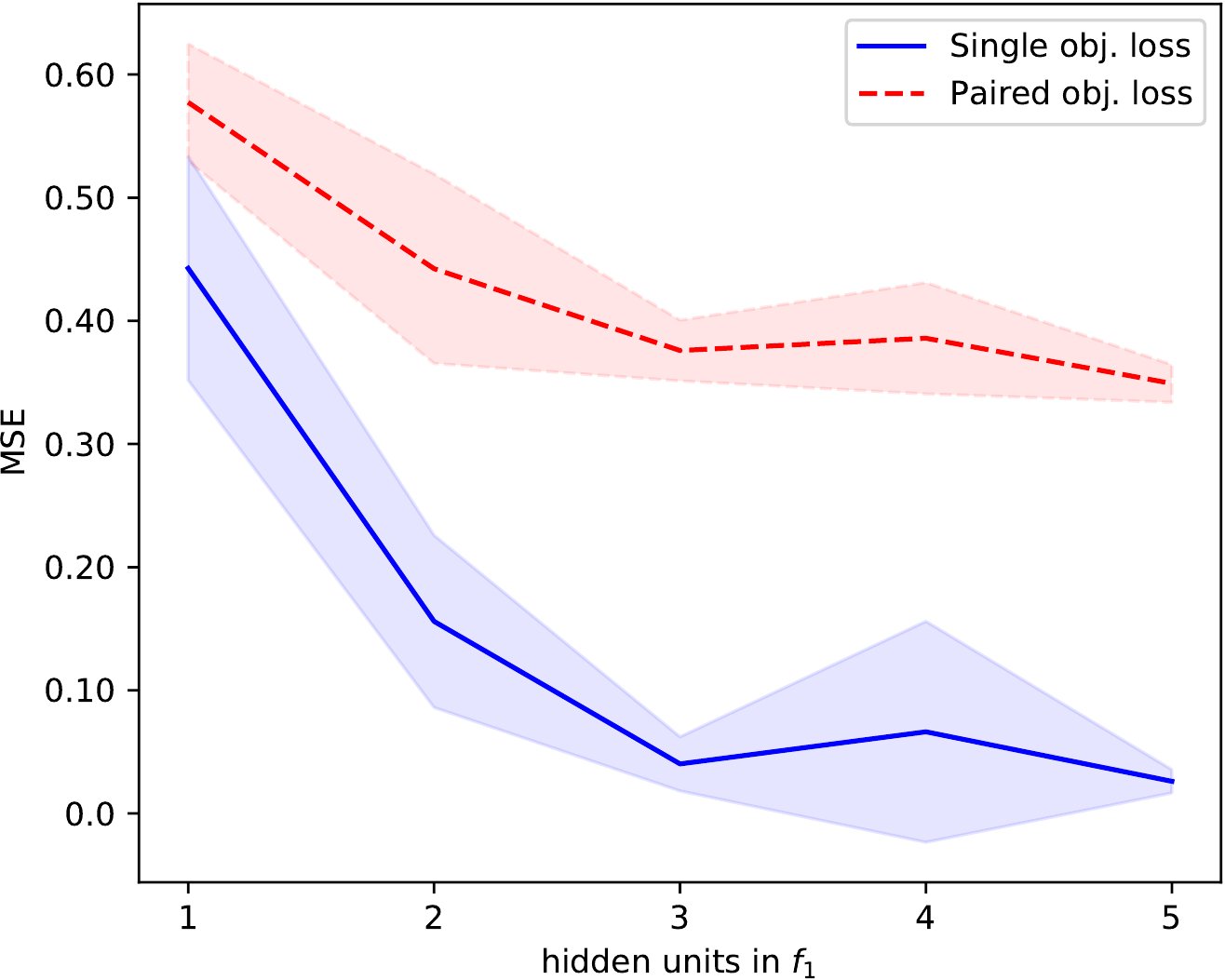}
        \caption{MSE vs. the number of units in the bottleneck of $f_1$.}
        \label{fig:mse1}
    \end{subfigure}
    \hfill
    \begin{subfigure}[b]{0.49\textwidth}
        \centering
        \includegraphics[width=\textwidth]{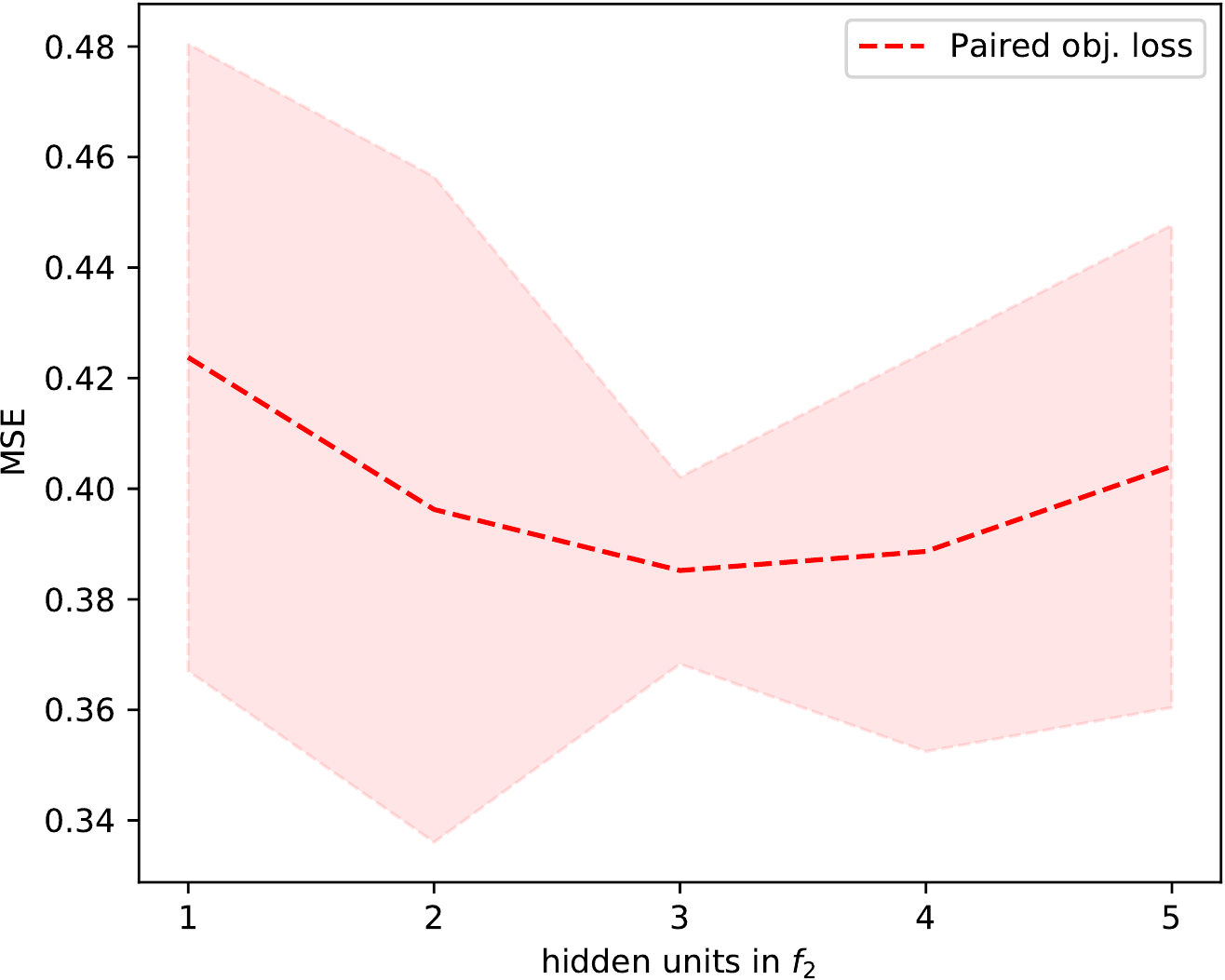}
        \caption{MSE vs. the number of units in the bottleneck of $f_2$.}
        \label{fig:mse2}
    \end{subfigure}
    \caption{The mean square error losses for $f_1$-$g_1$ and $f_2$-$g_2$ network pairs. In (a), we also plot the paired object MSE with a single unit for a varying number of units in the bottleneck of $f_1$ to show how different numbers of units affect MSE in new observations).}
    \label{fig:mse_curves}
\end{figure}

\subsection{MNIST 8-puzzle Environment}
Network architectures of the encoder and the decoder for the MNIST 8-puzzle environment are given in Tables \ref{tab:8puzzle_encoder} and \ref{tab:8puzzle_decoder}. For 8-puzzle w/r and 15-puzzle w/r versions, the bottleneck size is changed from 13 to 14 and 15, respectively. To ensure the output size for the 15-puzzle, we change the padding of the third and the fourth convolutional layer in the decoder from 1 to 0.

\begin{table}[htbp]
    \centering
    \begin{tabular}{|c|c|c|c|c|}
        \hline
        Layer & In ch. & Out ch. & Stride & Pad \\
        \hline
        \hline
        Conv4x4 & 1 & 64 & 2 & 1 \\
        \hline
        Conv4x4 & 64 & 128 & 2 & 1 \\
        \hline
        Conv4x4 & 128 & 256 & 2 & 1 \\
        \hline
        Conv4x4 & 256 & 512 & 2 & 1 \\
        \hline
        \multicolumn{5}{|c|}{Global average pooling over channels}\\
        \hline
        FC & 512 & 13 & - & - \\
        \hline
        \multicolumn{5}{|c|}{Gumbel-sigmoid}\\
        \hline
        \hline
        \multicolumn{5}{|c|}{Number of parameters: 2,763,085}\\
        \hline
    \end{tabular}
    \caption{The encoder for the 8-puzzle environment.}
    \label{tab:8puzzle_encoder}
\end{table}

\begin{table}[htbp]
    \centering
    \begin{tabular}{|c|c|c|c|c|}
        \hline
        Layer & In ch. & Out ch. & Stride & Pad \\
        \hline
        \hline
        FC & 13+4 & 512 & - & - \\
        \hline
        \multicolumn{5}{|c|}{Reshape (-1, 512) $\rightarrow$ (-1, 512, 1, 1)}\\
        \hline
        ConvT5x5 & 512 & 256 & 1 & 0 \\
        \hline
        ConvT4x4 & 256 & 128 & 2 & 1 \\
        \hline
        ConvT4x4 & 128 & 64 & 2 & 1 \\
        \hline
        ConvT4x4 & 64 & 32 & 2 & 1 \\
        \hline
        ConvT4x4 (no batch norm.) & 32 & 1 & 2 & 1 \\
        \hline
        \hline
        \multicolumn{5}{|c|}{Number of parameters: 3,976,097}\\
        \hline
    \end{tabular}
    \caption{The decoder for the 8-puzzle environment. ConvT stands for transposed convolutional layers. The last layer of the decoder does not include a batch normalization layer.}
    \label{tab:8puzzle_decoder}
\end{table}

\section{Using the Straight-Through Estimator}\label{app:ste}
The experiment results with STE on the tabletop environment are reported in Table \ref{tab:singlecats_ste}.

\begin{table}[htbp]
 \centering
 \begin{tabular}{l|cccc}
 \hline
 \multicolumn{5}{c}{DeepSym with STE} \\
 \hline
 Category & (0, 0) & (0, 1) & (1, 0) & (1, 1) \\
 \hline
 Sphere & \textbf{92.9 $\pm$ 8.6} & 2.5 $\pm$ 4.4 & 3.2 $\pm$ 6.8 & 1.4 $\pm$ 2.3 \\
 Cube & 1.4 $\pm$ 3.1 & \textbf{92.9 $\pm$ 10.3} & 3.4 $\pm$ 5.6 & 2.3 $\pm$ 3.7 \\
 Vertical Cylinder & 1.9 $\pm$ 4.6 & \textbf{93.6 $\pm$ 5.4} & 1.8 $\pm$ 2.7 & 2.7 $\pm$ 3.1 \\
 Horizontal Cylinder & 15.8 $\pm$ 27.3 & 7.6 $\pm$ 10.7 & \textbf{74.7 $\pm$ 27.0} & 2.0 $\pm$ 3.8\\
 Cup & 0.1 $\pm$ 0.2 & 0.0 $\pm$ 0.0 & 0.0 $\pm$ 0.0 & \textbf{99.9 $\pm$ 0.2}\\
 \hline
 \multicolumn{5}{c}{Autoencoder with STE (OBO)} \\
 \hline
 Category & (0, 0) & (0, 1) & (1, 0) & (1, 1) \\
 \hline
 Sphere & \textbf{85.1 $\pm$ 12.1} & 12.3 $\pm$ 9.6 & 2.6 $\pm$ 4.3 & 0.0 $\pm$ 0.0 \\
 Cube & \textbf{74.6 $\pm$ 17.7} & 20.4 $\pm$ 12.0 & 5.0 $\pm$ 7.0 & 0.0 $\pm$ 0.0 \\
 Vertical Cylinder & \textbf{76.2 $\pm$ 15.6} & 18.3 $\pm$ 9.4 & 5.5 $\pm$ 8.5 & 0.0 $\pm$ 0.0 \\
 Horizontal Cylinder & \textbf{78.1 $\pm$ 14.4} & 19.2 $\pm$ 11.1 & 2.8 $\pm$ 3.8 & 0.0 $\pm$ 0.0 \\
 Cup & \textbf{92.4 $\pm$ 14.1} & 6.6 $\pm$ 13.5 & 0.9 $\pm$ 2.8 & 0.1 $\pm$ 0.2 \\
 \hline
 \end{tabular}
 \caption{The relative assignment frequencies of objects to different symbolic categories. Here, objects vary in their sizes and initial positions. The mean and the standard deviation of 10 runs are reported.}
 \label{tab:singlecats_ste}
\end{table}

\section{The Number of States in 8-puzzle w/r and 15-puzzle w/r}\label{app:num_states}
For the 8-puzzle w/r, the number of possible states increases from $9!=362880$ to $9\times9^8=387420489$, which is an increase by about a factor of 1000. In general, the number of states is $n^2 k^{(n^2-1)}$ where $n$ stands for the size of the board (the size is 3 for 8-puzzle and 4 for 15-puzzle), and $k$ is the number of possible digits other than the empty tile. This translates to $\approx 3.29\times10^{15}$ states for 15-puzzle w/r. On the other hand, the number of states that the encoder should represent is $(n-2)^2k^4 + 4(n-2)k^3 + 4k^2$, which translates to 9801 and 32400 states for 8-puzzle w/r and 15-puzzle w/r, respectively. Therefore, we train DeepSym with 14 units for 8-puzzle w/r ($\log_2 9801 \approx 13.26$) and with 15 units for 15-puzzle w/r ($\log_2 32400 \approx 14.98$). In a sense, we use the most strict limit for the number of units.

\section{Symbols learned in 8-puzzle w/r and 15-puzzle w/r}\label{app:more_symbols}

\begin{figure}[htbp]
    \centering
    \includegraphics[width=\textwidth]{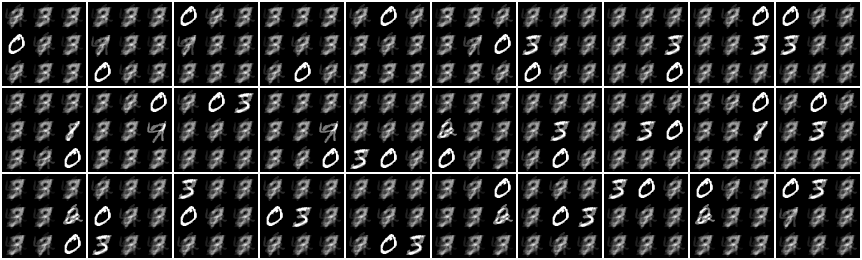}
    \caption{Average states that correspond to symbols learned in 8-puzzle w/r environment.}
    \label{fig:mnist_symbols_3r}
\end{figure}

\begin{figure}[htbp]
    \centering
    \includegraphics[width=\textwidth]{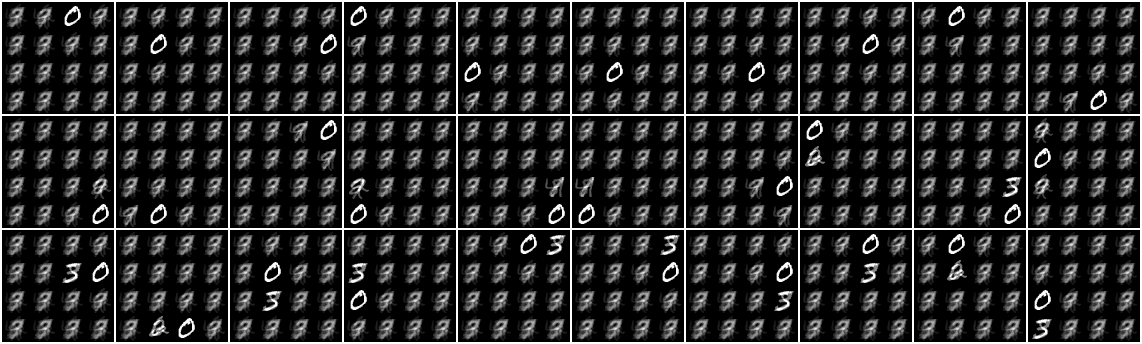}
    \caption{Average states that correspond to symbols learned in 15-puzzle w/r environment.}
    \label{fig:mnist_symbols_4r}
\end{figure}

\begin{figure}[htbp]
    \centering
    \includegraphics[width=\textwidth]{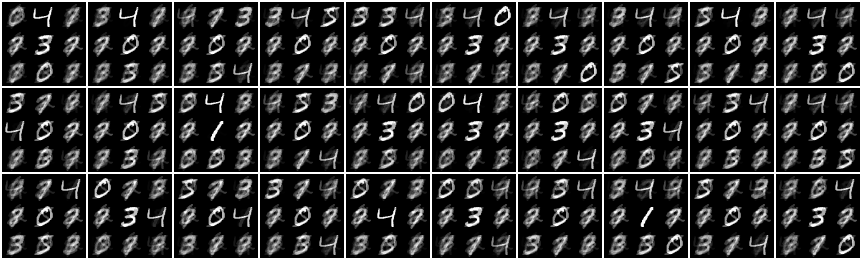}
    \caption{Average states that correspond to autoencoder symbols learned in the 8-puzzle environment.}
    \label{fig:sae_3p}
\end{figure}

\begin{figure}[htbp]
    \centering
    \includegraphics[width=\textwidth]{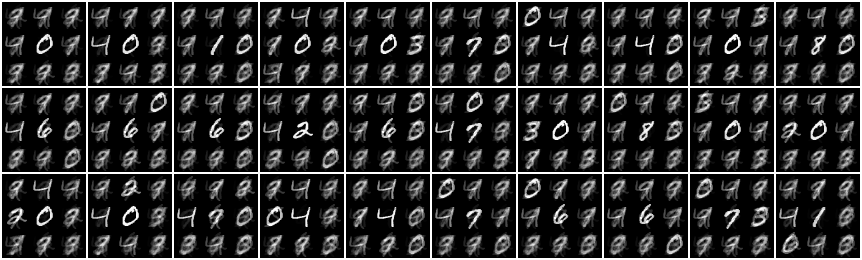}
    \caption{Average states that correspond to autoencoder symbols learned in 8-puzzle w/r environment.}
    \label{fig:sae_3r}
\end{figure}

\begin{figure}[htbp]
    \centering
    \includegraphics[width=\textwidth]{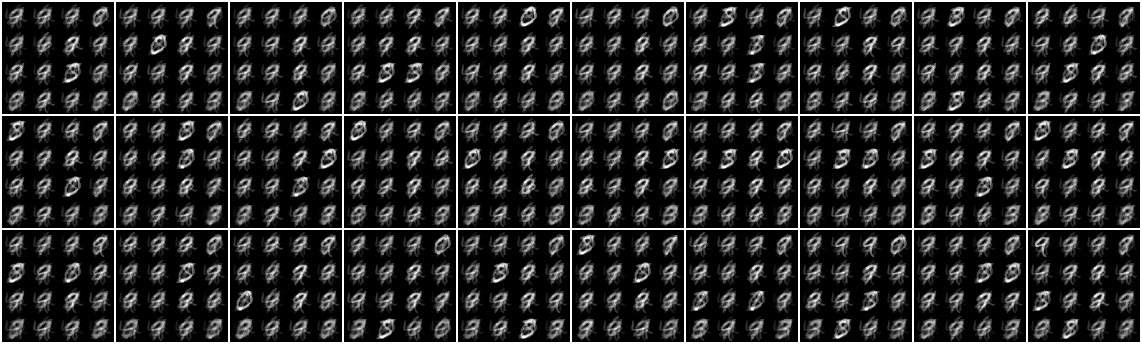}
    \caption{Average states that correspond to autoencoder symbols learned in 15-puzzle w/r environment.}
    \label{fig:sae_4r}
\end{figure}

\newpage

\vskip 0.2in
\bibliographystyle{theapa}
\bibliography{references}

\end{document}